\documentclass{article}
\usepackage{amssymb}

\usepackage[normalem]{ulem}
\usepackage{amsmath}
\usepackage[table]{xcolor} 
\usepackage{changepage,threeparttable}
\usepackage{graphicx,xcolor}
\usepackage{placeins}

\usepackage{subcaption}
\usepackage{booktabs}

\usepackage[normalem]{ulem}
\usepackage{color}

\usepackage{amssymb,amsmath,array}
\usepackage{algorithmicx}
\usepackage{algpseudocode}
\usepackage{float}
\usepackage[linesnumbered,ruled,vlined]{algorithm2e}
\usepackage{graphicx}
\usepackage{enumerate}
\usepackage[shortlabels]{enumitem}

\usepackage{arydshln}
\usepackage{tikz}
\usetikzlibrary{positioning, shapes, shadows, arrows}
\usetikzlibrary{decorations.pathreplacing}
\tikzstyle{abstract}=[circle, draw=black, fill=white]
\tikzstyle{labelnode}=[circle, draw=white,opacity=.2,text opacity=1]
\tikzstyle{invisiblenode}=[circle,dashed, inner sep=1pt,circle split,line width=1mm,minimum size=1.5cm]
\tikzstyle{line} = [draw, -latex']

\let\vec\mathbf
\usepackage[normalem]{ulem} 

\usepackage[utf8]{inputenc}

\title{Exploiting auto-encoders and segmentation methods for middle-level explanations of image classification systems}
\author{Andrea~Apicella,
        Salvatore~Giugliano,
        Francesco~Isgrò,\\
        and Roberto~Prevete\\
        \small Laboratory of Augmented Reality for Health Monitoring (ARHeMLab),\\
        \small Laboratory of Artificial Intelligence, Privacy \& Applications (AIPA Lab),\\
        \small Department of Electrical Engineering and Information Technology,\\ \small University of Naples Federico II \\
        \small Corresponding author: Andrea Apicella, and.api.univ@gmail.com
        }

\date{\footnote{This is a preprint of the work published in \textit{Knowledge-Based Systems} \\DOI: \url{https://doi.org/10.1016/j.knosys.2022.109725}}}

\usepackage{graphicx}

\usepackage{hyperref}
\usepackage[export]{adjustbox}
\usepackage[font=small,labelfont=bf]{caption}
\usepackage{lineno}

\newcommand{\nonl}{\renewcommand{\nl}{\let\nl\oldnl}}
\begin{document}

\maketitle

\begin{abstract}
A central issue addressed by the rapidly growing research area of eXplainable Artificial Intelligence (XAI) is to provide methods to give explanations for the behaviours of Machine Learning (ML) non-interpretable models after the training.
Recently, it is becoming more and more evident that new directions to create better explanations should take into account what a good explanation is to a human user. This paper suggests taking advantage of developing an XAI framework that allows producing multiple explanations for the response of image a classification system  in terms of potentially different middle-level input features.
To this end, we propose an XAI framework able to construct explanations in terms of input features extracted by auto-encoders. We start from the hypothesis that some auto-encoders, relying on standard data representation approaches, could extract more salient and understandable input properties, which we call here \textit{Middle-Level input Features} (MLFs), for a user with respect to raw low-level features. Furthermore, extracting different types of MLFs through different type of auto-encoders, different types of explanations for the same ML system behaviour can be returned. We experimentally tested our method on two different image datasets and using three different types of MLFs. The results are encouraging. Although our novel approach was tested in the context of image classification, it can potentially be used on other data types to the extent that auto-encoders to extract humanly understandable representations can be applied.
\end{abstract}
\section{Introduction}
\label{sec:introduction}

\begin{figure}
    \centering
    \includegraphics[width=0.9\textwidth]{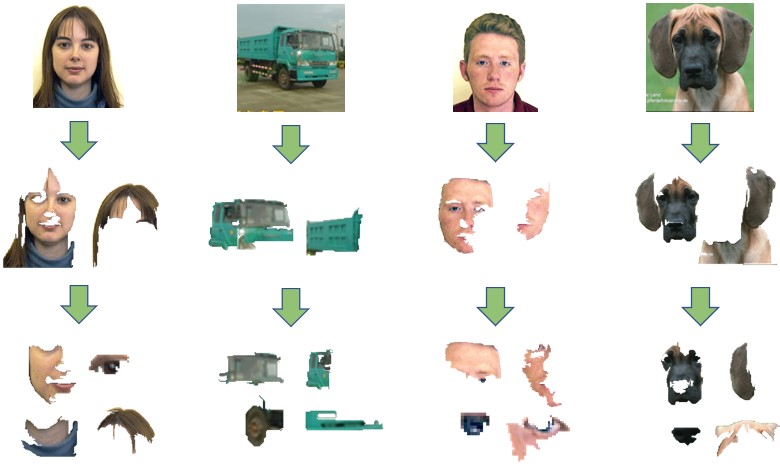}
    \caption{Examples of Middle Level input Features (MLFs). Each MLF represents a part of the input which is perceptually and cognitively salient to a human being, as for example the ears of a cat or the wings of an airplane. These features are intuitively more humanly interpretable respect to low-level features (as for example raw unrelated image pixels), so a decision explanation expressed in terms of MLF relevance can be easier to understand for a human being respect to explanations expressed in terms of low level features.
    }
    \label{fig:esempi_mlf}
\end{figure}

A large part of Machine Learning(ML) techniques – including Support Vector Machines (SVM) and Deep Neural Networks (DNN) – give rise to systems having behaviours often complex to interpret \cite{adadi2018}. More precisely, although ML techniques with reasonably well interpretable mechanisms and outputs exist, as, for example, decision trees, the most significant part of ML techniques give responses whose relationships with the input are often difficult to understand. In this sense, they are commonly considered as black-box systems. 
In particular, as ML systems are being used in more and more domains and, so,  by a more varied audience, there is the need
for making them understandable and trusting to general users \cite{ribera2019can,arrieta2020explainable}.
Hence, generating explanations for ML system behaviours that are understandable to human beings is a central scientific and technological issue addressed by the rapidly growing research area of eXplainable Artificial Intelligence (XAI). 
Several definitions of interpretability/explainability for ML systems have been discussed in the XAI literature \cite{doran2017,arrieta2020explainable}, and many approaches to the problem of overcoming their opaqueness are now
pursued \cite{nguyen2016_2,bach2015,apicella2020middle}.
For example, in \cite{montavon2018methods}
 a series of techniques for the interpretation of DNNs is discussed, and in \cite{lipton2018mythos} the authors examine and discuss the motivations underlying the interest in ML systems' interpretability, discussing and refining this notion. In the literature, particular attention is given to \textit{post-hoc} explainability \cite{arrieta2020explainable}, i.e., the methods to provide explanations for the behaviours of non-interpretable models after the training.
In the context of this multifaceted interpretability problem, we note that in the literature, one of the most successful strategies is to provide explanations in terms of "visualisations" \cite{ribera2019can,zhang2018}. More specifically, explanations for image classification systems are given in terms 
of low-level input features, such as relevance or heat maps of the input built by model-agnostic (without disclosing the model internal mechanisms) or model-specific (accessing to the  model internal mechanisms) methods, like sensitivity analysis\cite{simonyan2014} or Layer-wise Relevance Propagation (LRP) \cite{bach2015}. 
For example, LRP associates a relevance value to each input element (pixels in case of images) to explain the ML model answer. 
The main problem with such methods is that human users are left with a significant interpretive burden. 
Starting from each low-level feature's relevance, the human user needs to identify the overall input properties perceptually and cognitively salient to him \cite{apicella2020middle}. 
Thus, an XAI approach should alleviate this weakness of low-level approaches and overcome their limitations, allowing the possibility to construct explanations in terms of input features that represent more salient and understandable input properties for a user, which we call here Middle-Level input Features (MLFs) (see Figure \ref{fig:esempi_mlf}).
Although there is a recent research line which attempts to give explanations in terms of visual human-friendly concepts \cite{kim2018interpretability,ghorbani2019towards,akula2020cocox} (we will discuss them in Section \ref{sec:related_works}), however we notice that the goal to learn data representations that are easily factorised in terms of meaningful features is, in general, pursued in the \textit{representation learning} framework \cite{bengio2013representation}, and more recently in the \textit{feature disentanglement learning} context \cite{locatello2019challenging}.  These meaningful features may represent parts of the input such as nose, ears and paw in case of, for example, face recognition tasks (similarly to the outcome of a clustering algorithm) or more abstract input properties such as shape, viewpoint, thickness, and so on, leading to data representations perceptually and cognitively salient to the human being. 
Based on these considerations, in this paper, we propose to develop an XAI approach able to give explanations for an image classification system in terms of features which are obtained by standard representation learning methods such as variational auto-encoder \cite{chen2018isolating} and hierarchical image segmentation \cite{galvao2020image}. In particular, we exploit middle-level data representations obtained by auto-encoder methods \cite{charte2020analysis} to provide  explanations of image classification systems.
In this context, in an earlier work \cite{apicella2021explanations} we proposed an initial experimental investigation on this type of explanations exploiting the \textit{hierarchical} organisation of the data in terms of more elementary factors. For example, natural images can be described in terms of the objects they show at various levels of granularity \cite{tschannen2018recent,sonderby2016ladder,zhao2017learning}. Or in \cite{gu2019hierarchical} a hierarchical prototype-based approach for classification is proposed. This method has a certain degree of intrinsic transparency, but it does not fall into post-hoc explainability  category. 

To the best of our knowledge, in the XAI literature, however, there are relatively few approaches that pursue this line of research. 
In \cite{ribeiro2016}, the authors proposed LIME, a successful XAI method which is based, in case of image classification problems, on explanations expressed as sets of regions, clusters of the image, said superpixels which are obtained by a clustering algorithm. 
These superpixels can be interpreted as MLFs.
In \cite{apicella2020middle,Apicella2019d} the explanations are formed of elements selected from a dictionary of MLFs, obtained by sparse dictionary learning methods \cite{donnarumma2021framework}. 
In \cite{guidotti2020explaining} authors propose to exploit the latent representations learned through an adversarial auto-encoder for generating a synthetic neighbourhood of the image for which an explanation is required.
However, these approaches propose specific solutions which cannot be generalised to different types of input properties. By contrast, in this paper, we investigate the possibility of obtaining explanations using an approach that can be applied to different types of MLFs, which we call \textit{General MLF Explanations} (GMLF). 
More precisely, we develop an XAI framework that can be applied whenever a) the input of an ML system can be encoded and decoded based on MLFs, and b) any Explanation method producing a Relevance Map (ERM method) can be applied on both the ML model and the decoder. In this sense, we propose a general framework insofar as it can be applied to several different computational definitions of MLFs and a large class of ML models. Consequently, we can provide multiple and different explanations based on different MLFs. In particular, in this work we tested our novel approach in the context of image classification using MLFs extracted by three different methods: 1) image segmentation by auto-encoders, 2) hierarchical image segmentation by auto-encoders, and 3) Variational auto-encoders.  About the points 1) and 2), a simple method to represent the output of a  segmentation algorithm in terms of encoder-decoder is reported. However, this approach can be used on a wide range of different data types to the extent that encoder-decoder methods can be applied.

Thus, the medium or long-term objective of this research work is to develop a XAI general approach producing explanations for an ML System behaviour in terms of potentially different and user-selected input features,  composed of input properties which the human user can select according to his background knowledge and goals. This aspect can play a key role in developing \textit{user-centred} explanations. 
It is essential to note that, in making an explanation understandable for a user, it should be taken into account what information the user desires to receive \cite{kim2018interpretability,ribera2019can,Apicella2019b,Apicella2019c}. Recently, it is becoming more and more evident that new directions to create better explanations should take into account what a good explanation is for a human user, and consequently to develop XAI solutions able to provide user-centred explanations \cite{kim2018interpretability,ribera2019can,lim2019these,akula2020cocox,kim2014advertiser}. By contrast, much of the current XAI methods provide specific ways to build explanations that are based on the researchers' intuition of what constitutes a "good" explanation \cite{lim2019these,miller2019explanation}.

\color{black}
To summarise, in this paper the following novelties are presented:
\begin{enumerate}
\item a XAI framework where middle-level or high-level input properties can be built exploiting standard methods of data representation learning is proposed;
\item our framework can be applied  to several different computational definitions of middle-level or high-level input properties and a large class of ML models. Consequently, multiple and different explanations based on different middle-level input properties can be possibly provided given an input-ML system response;
\item The middle-level or high-level input proprieties are computed independently from the ML classifier to be explained.
\end{enumerate}
\color{black}

The  paper  is  organised  as  follows:  Section  \ref{sec:method} describes in detail the  proposed approach; in Section \ref{sec:related_works} we discuss  differences and advantages of GMLF with respect similar approaches presented in the literature; experiments and results are discussed in Section \ref{sec:results}. In particular, we compared our approach with LIME method and performed both qualitative and quantitative evaluations of the results; the concluding Section summarises the main high-level features  of  the  proposed  explanation framework and outlines some future developments.
\section{Related Works}
\label{sec:related_works}
The importance of eXplainable Artificial Intelligence (XAI) is discussed in several papers \cite{weller2017,miller2019explanation,samek2019,arrieta2019explainable}. Different strategies have been proposed to face the explainability problem, depending both on the AI system to explain and the type of explanation proposed. 
Among all the XAI works proposed over the last years, an important distinction is between model-based and post-hoc explanaibility \cite{murdoch2019definitions}, the former consisting in AI systems explainable by design (e.g., decision trees), since their inner mechanisms are easily interpreted, the latter proposing explanation built for system that are not easy to understand. 
In particular, several methods to explain Deep Neural Networks (DNNs) are proposed in the literature due to the high complexity of their inner structures. A very common approach consists in returning visual-based explanations in terms of input feature importance scores, as for example Activation Maximization (AM) \cite{erhan2009}, Layer-Wise Relevance propagation (LRP) \cite{bach2015}, Deep Taylor Decomposition\cite{binder2016,montavon2017_2}, Class Activation Mapping (CAM) methods \cite{zhou2016learning, selvaraju2017grad}, Deconvolutional Network \cite{zeiler2011} and \textit{Up-convolutional network} \cite{zeiler2014,dosovitskiy2016}. Although heatmaps seem to be a type of explanation that is easy to understand for the user, these methods build relevances on the low-level input features (the single pixel), while input middle-level properties which determined the answer of the classifier have to be located and interpreted by the user, leaving much of the interpretive work to the human beings.
On the other side, methods as Local Interpretable Model-agnostic Explanations (LIME) \cite{ribeiro2016} relies on feature partitions, as super-pixel in the image case. However, the explanations given by LIME (or its variants) are built through a new model that approximates the original one, thus risking to loose the real reasons behind the behaviour of the original model  \cite{ribera2019can}.

Recently, a growing number of studies \cite{zhou2018interpretable,kim2018interpretability,ghorbani2019towards,akula2020cocox} have focused on  providing explanations in the form of middle-level or high-level human ``concepts'' as we are addressing it in this paper. 

In particular, in \cite{kim2018interpretability} the authors introduce the Concept Activation Vectors (CAV) as a way of visually representing the neural network' inner states associated with a given class. CAVs should represent human-friendly concepts. The basic ideas can be described as follow: firstly, the authors suppose the availability of an external labelled dataset $XC$ where each label corresponds to a human-friendly concept. Then, given a pre-trained neural network classifier to be explained, say $NC$, they consider the functional mapping $f_l$ from the input to the $l$-layer of $NC$. Based on $f_l$, for each class $c$ of the dataset $XC$, they build a linear classifier composed of $f_l$ followed by a linear classifier to distinguish the element of $XC$ belonging to the class $c$ from randomly chosen images.  The normal to the learned hyperplane is considered the CAV for the user-defined concept corresponding to the class $c$. Finally, given all the input belonging to a class $K$ of the pre-trained classifier $NC$, the authors define a way to quantify how much a concept $c$, expressed by a CAV, influences the behaviour of the classifier, using directional derivatives to computes $NC$ 's conceptual sensitivity across entire class $K$ of inputs.

Building upon the paper discussed above, in \cite{akula2020cocox} the authors provide explanations in terms
of \textit{fault-lines}\cite{kahneman1981simulation}. Fault-lines should represent ``high-level semantic aspects
of reality on which humans zoom in when imagining an alternative to it''.
Each fault-line is represented by a minimal set of semantic \textit{xconcepts} that need to be added to or deleted from the classifier's input to alter the class that the classifier outputs.
Xconcepts are built following the method proposed in \cite{kim2018interpretability}. In a nutshell, given a pre-trained convolutional neural network $CN$ whose behaviour is to be explained, xconcepts are defined in terms of super-pixels (images or parts of images) related to the feature maps of the $l$-th $CN$'s convolutional layer, usually the last convolutional layer before the full-connected layer. In particular, these super-pixels are collected when the input representations at the convolution layer $l$ are used to discriminate between a target class $c$ and an alternate class $c_{alt}$, and they are computed based on the Grad-CAM algorithm \cite{selvaraju2017grad}. In this way, one obtains xconcepts in terms of images related to the class $c$ and able to distinguish it from the class $c_{alt}$.  
Thus, when the classifier $CN$ responds that an input $x$ belongs to a class $c$, the authors provide an explanation in terms of xconcepts which should represent semantic aspects of why $x$ belongs $c$ instead of an alternate class $c_{alt}$.  

In \cite{ghorbani2019towards} the authors propose a method to provide explanations related to an entire class of a trained neural classifier. The method is based on the CAVs introduced in \cite{kim2018interpretability} and sketched above. However, in this case, the CAVs are automatically extracted without the need an external labelled dataset expressing human-friendly concepts.

Many of the approaches discussed so far focus on \textit{global} explanations, i.e., explanations related to en entire class of the trained neural network classifier (see \cite{ghorbani2019towards,kim2018interpretability}). Instead, in our approach, we are looking for \textit{local} explanations, i.e., explanations for the response of the ML model to each single input. Some authors, see for example \cite{kim2018interpretability}, provide methods to obtain \textit{local} explanations, but in this case, the explanations are expressed in terms of high-level visual concepts which do not necessarily belong to the input. Thus, again human users are left with a significant interpretive load: starting from external high-level visual concepts, the human user needs to identify the input properties perceptually and cognitively related to these concepts. On the contrary, the input (MLFs) high-level properties are expressed, in our approach, in terms of elements of the input itself.

Another critical point is that high-level or middle-level user-friendly concepts are computed on the basis of the neural network classifier to be explained. In this way, a short-circuit can be created in which the visual concepts used to explain the classifier are closely related to the classifier itself. By contrast, in our approach, MLFs are extracted independently from the classifier.

A crucial aspect that distinguishes our proposal from the above-discussed research line is grounded on the fact that we propose an XAI framework able to provide multiple explanations, each one composed of a specific type of middle-level input features (MLFs). Our methodology only needs that MLFs can be obtained using methods framed into data representation research, and, in particular, any auto-encoder architecture for which an explanation method producing a relevance map can be applied on the decoder (see Section \ref{sec:method:general_description}).

To summarise, our GMLF approach, although shares with the above describe research works the idea to obtain explanations based on middle-level or high-level human-friendly concepts, presents the following elements of novelty:
\begin{enumerate}
\item It is a XAI framework where middle-level or high-level input properties can be built on the basis of standard methods of data representation learning.
\item It outputs local explanations.
\item The middle-level or high-level input proprieties are computed independently from the ML classifier to be explained.  
\end{enumerate}

Regarding points 2) and 3) we notice that a XAI method that has significant similarity with our approach is LIME \cite{ribeiro2016} or its variants (see, for example, \cite{zhao2020baylime}). LIME,  especially in the context of images,  is one of the predominant XAI methods discussed in the literature \cite{dieber2020model,zhao2020baylime}. It can provide \textit{local} explanations in terms of superpixels which are regions or parts of the input that the classifier receives, as we have already discussed in Section \ref{sec:introduction}. These superpixels can be interpreted as middle-level input properties, which can be more understandable for a human user than low-level features such as pixels. In this sense, we view a similarity in the output between our approach GMLF and LIME. The explanations built by LIME can be considered comparable with our proposed approach but different in the construction process. While LIME builds explanation relying on a proxy model different from the model to explain, the proposed approach relies only on the model to explain, without needing any other model that approximates the original one. To highlight the difference between the produced explanations, in section \ref{sec:exp} a comparison between LIME and GMLF outputs is made. 

\section{Approach}
\label{sec:method}
Our approach stems from the following observations.

The development of data representations from raw low-level data usually aims to obtain distinctive explanatory features of the data, which are more conducive to subsequent data analysis and interpretation.
This critical step has been tackled for a long time using specific methods developed exploiting expert domain knowledge. 
However, this type of approach can lead to unsuccessful results and requires a lot of heuristic experience and complex manual design \cite{li2020network}. 
This aspect is similar to what commonly occurs in many XAI approaches, where the explanatory methods are based on the researchers' intuition of what constitutes a "good" explanation.  

By contrast, representation learning successfully investigates ways to obtain middle/high-level abstract feature representations by automatic machine learning approaches. 
In particular, a large part of these approaches is based on  Auto-Encoder (AE) architectures \cite{charte2020analysis,li2020network}. 
AEs correspond to neural networks composed of at least one hidden layer and logically divided into two components, an \textit{encoder} and a \textit{decoder}.
From a functional point of view, an AE can be seen as the composition of two functions $E$ and $D$: $E$ is an encoding function  (the encoder) which maps the input space onto a feature space (or latent encoding space), $D$ is a decoding function (the decoder) which inversely maps the feature space on the input space. 
A meaningful aspect is that by AEs, one can obtain data representations in terms of latent encodings $\vec{h}$, where each $h_i$ may represent a MLF $\xi_i$ of the input , such as parts of the input (for example, nose, ears and paw) or more abstract features which can be more salient and understandable input properties for a user.
See for example variational AE \cite{kingma2013auto,rezende2015variational,li2019disentangled} or image segmentation \cite{gao2016efficient,chen2018encoder,yu2018unsupervised,zhang2021improved} (see Figure \ref{fig:esempi_mlf}). 
Furthermore, different AEs can extract different data representations which are not mutually exclusive.   

Based on the previous considerations, we want to build upon the idea that the elements composing an explanation can be determined by an AE which extracts relevant input features for a human being, i.e., MLFs, and that one might change the type of MLFs changing the type of auto-encoder or obtain multiple and different explanations based on different MLFs.    

\subsection{General description}
\label{sec:method:general_description}
Given an ML classification model $M$ which receives an input $\vec{x} \in R^d$ and outputs $\vec{y} \in R^c$,
our approach can be divided into two consecutive steps. 

In the first step, we build an auto-encoder $AE \equiv (E,D)$ such that each input $\vec{x} $ can be encoded by $E$ in a latent encoding $ \vec{h} \in R^m$ and decoded by $D$. 
As discussed above, to each value $h_i$ is associated a MLF $\xi_j$ 
, thus each input $x$ is decomposed in a set of $m$ MLFs $\vec{\xi}=\{\xi_i\}_{i=1}^m$, where to each $\xi_i$ is associated the value $h_i$.
Different choices of the auto-encoder can lead to MLFs $\vec{\xi_i}$ of different nature, so to highlight this dependence we re-formalise this first step as follows: we build an encoder $E_{\vec{\xi}}:\vec{x} \in R^d \rightarrow \vec{h} \in R^m$ and a decoder $D_{\vec{\xi}}:\vec{h} \in R^m \rightarrow \vec{x} \in R^d$, where $\vec{h}$ encodes $\vec{x}$ in terms of the MLFs $\vec{\xi}$. 

\begin{figure}
    \centering
    \includegraphics[width=0.65\textwidth]{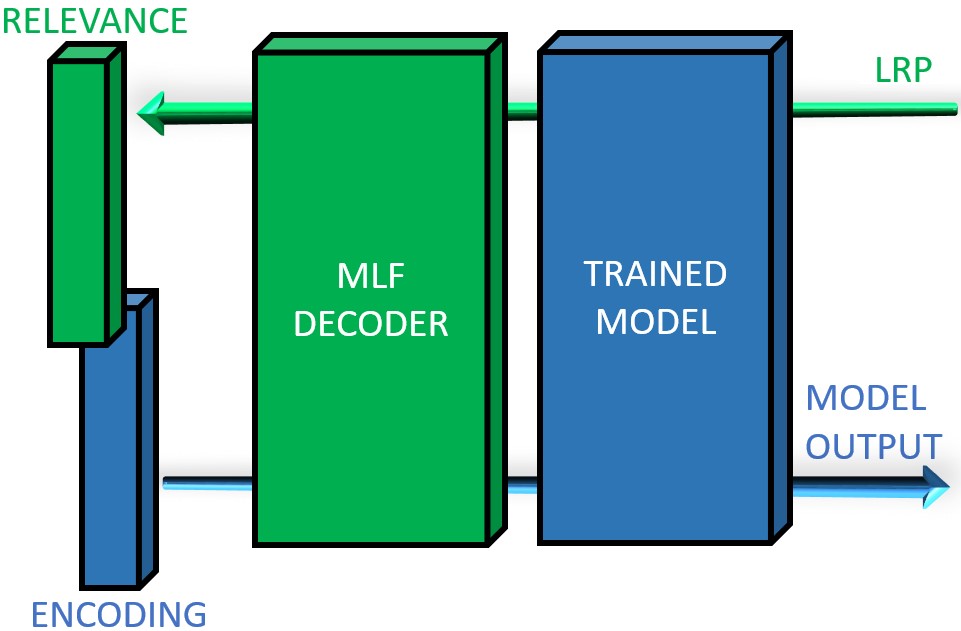}
    \caption{A general scheme of the proposed explanation framework. Given a middle-level feature encoder and the respective decoder, this last one is stacked on the top of the model to inspect. Next, 
    the encoding of the input is fed to the decoder-model system. A backward relevance propagation algorithm is then applied.}
    \label{fig:arch_1}
\end{figure}

In the second step of our approach, we use an ERM method (an explanation method producing a relevance map of the input) on both $M$ and $D_{\vec{\xi}}$, i.e., we apply it on the model $M$ and then use the obtained relevance values to apply the ERM method on $D_{\vec{\xi}}$  getting a relevance value for each middle-level feature.
In other words, we stack $D_{\vec{\xi}}$ on the top of $M$ thus obtaining a new model $DM_{\vec{\xi}}$ which receives as input an encoding $\vec{u}$ and outputs $\vec{y}$, and uses an ERM method on $DM_{\vec{\xi}}$ from $\vec{y}$ to $\vec{u}$. 
 In Figure \ref{fig:arch_1} we give a graphic description of our approach GMLF, and in algorithm \ref{algo:method_main}) it is described in more details considering a generic auto-encoder, while in algorithms \ref{algo:hier2} and  \ref{algo:VAE} our approach (GMLF) is described in case of specific auto-encoders (see Section  \ref{sec:method:hierarchical} and \ref{sec:method:vae}).
 
Thus, we search for a relevance vector $\vec{u}\in R^m$ which informs the user how much each MLF of $\vec{\xi}$ has contributed to the ML model answer $\vec{y}$. 
Note that, GMLF can be generalised to any decoder $D_{\vec{\xi}}$ to which a ERM method applies on. In this way, one can build different explanations for a $M$'s response in terms of different MLFs $\vec{\xi}$.  

In the remainder of this section, we will describe three alternative ways (segmentation, hierarchical segmentation and VAE) to obtain a decoder such that a ERM method can be applied to, and so three ways of applying our approach GMLF. We experimentally tested our framework using all the methods.  

\begin{algorithm}[t!]
\SetAlgoLined

\KwIn{data point $\vec{x}$, trained model $M$, an ERP method $RP$}
\KwOut{Feature Relevances $U$}
        $\vec{y} \leftarrow M(\vec{x})$\;
        build an auto-encoder $AE \equiv (E_{\xi},D_{\xi})$\;
        $\vec{h} \leftarrow E_{\xi}(\vec{x})$\;
        define $R: \vec{h}\mapsto \vec{x}-D_{\xi}(\vec{h})$\;
        define $DM_{\xi} : \vec{h} \mapsto M (D_{\xi}(\vec{h}) + R(\vec{h}))$ \;
        $U\leftarrow RP(DM_{\xi},\vec{h}, \vec{y})$  \;
		\textbf{\textbf{return} $U$}\;
 \caption{Proposed method GMLF}
 \label{algo:method_main}
\end{algorithm}

\subsection{MLFs from image segmentation} 
\label{sec:method:hierarchical}
\begin{figure}
    \centering
    \includegraphics[width=1\textwidth]{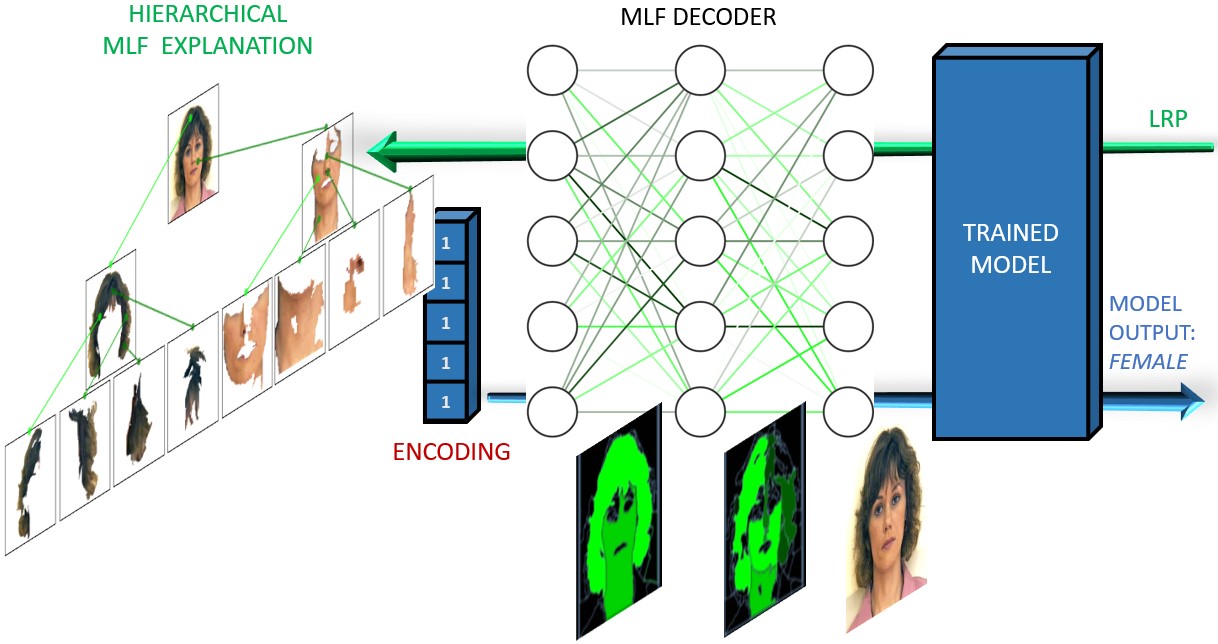}
    \caption{A segmentation-based MLF framework. MLF decoder is built as a neural network having as weights the segments returned by a hierarchical segmentation algorithm (see text for further details). The initial encoding is the "1" vector since all the segments are used to compose di input image. The relevance backward algorithm returns the most relevant segments.}
    \label{fig:arch_hier}
\end{figure}

Here we describe the implementation of the GMLF approach to the case of an auto-encoder built of the basis of hierarchical segmentation.
The approach is depicted in Figure \ref{fig:arch_hier}, while we give an algorithmic formalisation in algorithms \ref{algo:hier} and \ref{algo:hier2}.

Given an image $\vec{x}\in R^d$, a segmentation algorithm returns a partition of $\vec{x}$ composed of $m$ regions $\{q_i\}_{i=1}^m$. Some of the existing segmentation algorithms can be considered \textit{hierarchical segmentation algorithms}, since they return partitions hierarchically organised with increasingly finer levels of details.

More precisely, following \cite{guimaraes2012hierarchical}, we consider a segmentation algorithm to be hierarchical if it ensures both the causality principle of multi-scale analysis \cite{guigues2006scale} (that is, if a contour is present at a given scale, this contour has to be present at any finer scale) and the location principle (that is, even when the number of regions decreases, contours are stable). 
These two principles ensure that the segmentation obtained at a coarser detail level can be obtained by merging regions obtained at finer segmentation levels.

In general, given an image, a possible set of MLFs can be the result of a segmentation algorithm. Given an image $\vec{x} \in R^d$, and a partition of $\vec{x}$ consisting of $m$ regions $\{q_i\}_{i=1}^m$, each image's region $q_i$ can be represented by a vector $\vec{v}_i \in R^d$ defined as follows: $v_{ij}=0$ if $x_j \notin q_i$, otherwise $v_{ij}=x_j$, and $\sum_{i=1}^m \vec{v_i} =\vec{x}$. 
Henceforth, for simplicity and without loss of generality, we will use $\vec{v}_i$ instead of $q_i$ since they represent the same entities. 
Consequently, $\vec{x}$ can be expressed as linear combination of the $\vec{v}_i$ with all the coefficients equal to $1$, which represent the encoding of the image $\vec{x}$ on the basis of the $m$ regions. 
More in general, given a set of $K$ different segmentations $\{S_1,S_2,\dots,S_K\}$ of the same image sorted from the coarser to the finer detail level, it follows that, if the segmentations have a hierarchical relation, each coarser segmentation can be expressed in terms of the finer ones. 
More in detail, each region $\vec{v}^k_i$ of $S_k$ can be expressed as a linear combination $\sum_j\alpha_j \vec{v}^{k+1}_j$ where $\alpha_j$ is $1$ if all the pixels in $\vec{v}^{k+1}_j$ belong to $\vec{v}^{k}_i$, $0$ otherwise. 
We can apply the same reasoning going from $S_K$ to the image $\vec{x}$ considering it as a trivial partition $S_{K+1}$ where each region represents a single image pixel, i.e., $S_{K+1}=\{\vec{v}_1^{K+1},\vec{v}_2^{K+1},\dots, \vec{v}_d^{K+1}\}$, with $v^{K+1}_{ij}=x_j$ if $i=j$, otherwise $v^{K+1}_{ij}=0$ . 

It is straightforward to construct a feed-forward full connected neural network of $K+1$ layers representing an image $\vec{x}$ in terms of a set of $K$ hierarchically organised segmentations $\{S_k\}_{k=1}^K$ as follows (see Figure \ref{fig:arch_hier}): the $k$-th network layer has $|S_k|$ inputs and $|S_{k+1}|$ outputs, the identity as activation functions, biases equal to $0$ and each weights $w^k_{ij}$ equal to $1$ if the $\vec{v}^{k+1}_j$ region belongs to the $\vec{v}^{k}_i$ region, $0$ otherwise. The last layer $K+1$ has $d$ outputs and weights equal to $(\vec{v}_p^{K+1})_{p=1}^d$. The resulting network can be viewed as a decoder that, fed with the $\vec{1}$ vector, outputs the image $\vec{x}$. 

Note that if one considers $K=1$, it is possible to use the same approach in order to obtain an $\vec{x}$'s segmentation without a hierarchical organisation. In this case the corresponding decoder is a network composed of just one layer. 
\color{black}
We want to clarify that the segmentation module described in this section represents a way to build an auto-encoder which encodes latent variables that are associate to image segments. These image segments are candidate MLFs. Explanations are built by a selection of these candidate segments in the second computational step of our approach. We emphasize that the first step of our framework is to build an auto-encoder so that each input can be decomposed in a set of MLFs where each latent variable is associated to a specific MLF. These MLFs represent candidate input properties to be included into the final explanation which is computed by the second computational step of our approach. In this second part, a number of candidate middle-level input features are selected by an explanation method producing a relevance map of the input such as Layer-wise Relavance Propagation method (LRP). However,  different choices of the auto-encoder can lead to different MLFs of different nature.  
\color{black}

\begin{algorithm}[!]
\SetAlgoLined

\KwIn{data point $\vec{x}$, hierarchical segmentation procedure $seg$, hierarchical segmentation parameters $\vec{\lambda}=(\lambda_1,\lambda_2,\dots, \lambda_K)$}
\KwOut{ A Decoder $D_{\xi}$, an Encoder $E_{\xi}$}
$\{S_1,S_2,\dots,S_K\}\leftarrow seg(\vec{x},\vec{\lambda})$\;
$S_{K+1}\leftarrow \emptyset$\;
\For{ $x_j \in \vec{x}$}{
   let $\vec{v}^{K+1} \in \{0\}^d$\;
   $v^{K+1}_{jj}\leftarrow x_j$\;
   $S_{K+1}\leftarrow S_{K+1} \cup \{\vec{v}^{K+1}\}$\;
}

\For{$1\leq k \leq K$}{
 let $W^k \in \{0\}^{|S_k|\times|S_{k+1}|}$\;
 let $\vec{b}^k \in \{0\}^{|S_{k+1}|}$\;
 \For{$1\leq i \leq |S_k|$}{
    \For{$1\leq j \leq |S_{k+1}|$}{
         \If{$\vec{v}_j^{k+1}$ \textbf{belongs to} $\vec{v}_i^{k}$}{
            $W_{ij} \leftarrow 1$\;
         }
     }
 }
 define $identity: \vec{a} \mapsto \vec{a}$\;
 $D_{\xi}\leftarrow generateNeuralNetwork(\text{weights}=\{W^k\}_{k=1}^{K+1},$\\$\hspace{168pt}\text{biases}=\{\vec{b}^k\}_{k=1}^{K+1},$\\$\hspace{168pt}\text{activation function}=identity)$\;
 define $E_{\xi}: \vec{x} \mapsto e \in \{1\}^{|S_1|} $\;
 \textbf{\textbf{return} $D_{\xi},E_{\xi}$}\;
}

 \caption{Hierarchical segmantion-based Enconder-Decoder Generator}
 \label{algo:hier}
\end{algorithm}

\begin{algorithm}[!]
\SetAlgoLined

\KwIn{a data point $\vec{x} \in R^{d}$, a $trainedNeuralNet$ returning the class scores given a data point, a hierarchical segmentation procedure $seg$, hierarchical segmentation parameters $\vec{\lambda}=(\lambda_1,\lambda_2,\dots, \lambda_K)$, a relevance propagation algorithm $RP$ returning a relevance vector for each network layer given: i) a neural network, ii) an input and iii) its class probabilities,
a $generateNeuralNetwork$ function that returns a neural networks with weights, biases and activation function given as parameters}
\KwOut{ relevances for the first $K$ layers $\{\vec{u}_1,\dots,\vec{u}_K\}$}
\nonl
\setlist{nolistsep}
\begin{enumerate}[label=\scriptsize\textbf{\arabic*}, noitemsep]
 \item $\vec{y} \leftarrow M(\vec{x})$\;
\setlist{nolistsep}
\begin{enumerate}[\scriptsize a),noitemsep]
    \item $\vec{y} \leftarrow TrainedNeuralNet(\vec{x})$\;
\end{enumerate}
\item build an auto-encoder $AE \equiv (E_{\xi},D_{\xi})$\;
\setlist{nolistsep}
\begin{enumerate}[\scriptsize a),noitemsep]
\item $(E_{\xi},D_{\xi}) \leftarrow buildAE(\vec{x}, seg, \vec{\lambda})$\algorithmiccomment{see algorithm \ref{algo:hier}}\;
\end{enumerate}
\item $\vec{h} \leftarrow E_{\xi}(\vec{x})$\;
\item define $R: \vec{h}\mapsto \vec{x}-D_{\xi}(\vec{h})$ :
\setlist{nolistsep}
\begin{enumerate}[\scriptsize a),noitemsep]
\item let $W_{res} \in \{0\}^{ d \times d}$\;
\item $\vec{r}=\vec{x}-D_{\xi}(\vec{x})$\;
\item $\vec{b}_{res} \leftarrow \vec{r}$\;
\item define $identity: \vec{a} \mapsto \vec{a}$\;
\item $R \leftarrow generateNeuralNetwork(\text{weights}=\{W_{res}\},$\\$\hspace{168pt}\text{biases}=\{\vec{b}_{res}\},$\\$\hspace{168pt}\text{activation function}=identity)$\;
\end{enumerate}
\item define $DM_{\xi} : \vec{h} \mapsto M (D_{\xi}(\vec{h}) + R(\vec{h}))$ :
\setlist{nolistsep}
\begin{enumerate}[\scriptsize a),noitemsep]
\item $DM_{\xi} \leftarrow$ stackTogether(D,R,M)\;
\end{enumerate}
\item $U \leftarrow RP(DM_{\xi}, \vec{h},\vec{y})$\;
\item return $\{\vec{u}_1,\dots \vec{u}_K\}$\;
\end{enumerate}

 \caption{GMLF approach in case of Hierarchical segmentation-based auto-encoder}
 \label{algo:hier2}
\end{algorithm}

\subsection{MLF from Variational auto-encoders}
\label{sec:method:vae}
The concept of ``entangled features'' is strictly related to the concept of ``interpretability''. 
As stated in \cite{higgins2017bvae}, a disentangled data representation is most likely more interpretable than a classical entangled data representation. 
This fact is due to the generative factors representation into separate latent variables representing single features of the data (for example, the size or the colour of the represented object in an image). 

Using Variational Auto Encoders (VAE) is one of the most affirmed neural network-based methods to generate disentangled encodings. In general, a VAE is composed of two parts. First, an encoder generates an entangled encoding of a given data point (in our case, an image). Then a decoder generates an image from an encoding.  
Once trained with a set of data,  the VAE output $\vec{\tilde x}$ on a given input $\vec{x}$  can be obtained as the composition of two functions, an encoding function $E(\cdot)$and a decoding function $D(\cdot)$, implemented as two stacked feed-forward neural networks. 

The encoding function generates a data representation $E(\vec{x}) = \vec{h}$ of an image $\vec{x}$, the decoding function generates an approximate version $D(\vec{h})=\vec{\tilde x}$ of $\vec{x}$ given the encoding $\vec{h}$, with a residual $\vec{r}=\vec{x}-\vec{\tilde x}$. 
So, it is possible to restore the original image data simply adding the residual to $\vec{\tilde x}$, that is $\vec{x}=\vec{\tilde x} + \vec{r}$. 
Consequently, we stack the decoder neural networks with a further dense layer $R(\cdot)$ having $d$ neurons with weights set to $0$ and biases set to $\vec{r}$. 
The resulting network $R(E(\vec{h}))$ generates $\vec{x}$ as output, given its latent encoding $\vec{h}$.  

\begin{figure}
    \centering
    \includegraphics[width=0.95\textwidth]{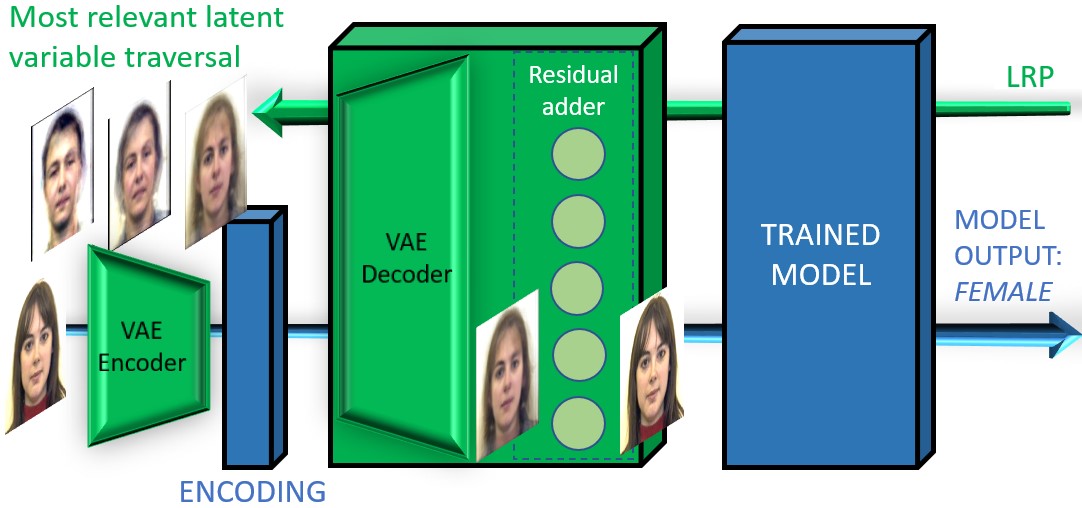}
    \caption{A VAE-based MLF framework. The MLF decoder is built as a neural network composed of the VAE decoder module followed by a full-connected layer containing the residual of the input (see text for further details). The initial input encoding is given by the VAE encoder module. The relevance backward algorithm returns the most relevant latent variables.}
    \label{fig:arch_vae}
\end{figure}

In Figure \ref{fig:arch_vae} it is shown a pictorial description of GMLF approach when the auto-encoder is built based on VAE, the algorithmic description is reported in algorithm \ref{algo:VAE}.

\begin{algorithm}[!]
\SetAlgoLined
\KwIn{a data point $\vec{x} \in R^{d}$, a $trainedNeuralNet$ returning the class scores given a data point, a $getTrainedVAE$ procedure returning a trained VAE, a relevance propagation algorithm $RP$ returning a relevance vector given: i) a neural network, ii) an input and iii) its class probabilities,
a $generateNeuralNetwork$ function returning a neural networks with weights, biases and activation function given as parameters}
\KwOut{ relevances $\vec{u}$ of each latent variable}
\nonl
\setlist{nolistsep}
\begin{enumerate}[label=\scriptsize\textbf{\arabic*}, noitemsep]
 \item $\vec{y} \leftarrow M(\vec{x})$\;
\setlist{nolistsep}
\begin{enumerate}[\scriptsize a),noitemsep]
    \item $\vec{y} \leftarrow TrainedNeuralNet(\vec{x})$\;
\end{enumerate}
\item build an auto-encoder $AE \equiv (E_{\xi},D_{\xi})$\;
\setlist{nolistsep}
\begin{enumerate}[\scriptsize a),noitemsep]
\item $(E_{\xi},D_{\xi}) \leftarrow getTrainedVAE()$\;
\end{enumerate}
\item $\vec{h} \leftarrow E_{\xi}(\vec{x})$\;
\item define $R: \vec{h}\mapsto \vec{x}-D_{\xi}(\vec{h})$ :
\setlist{nolistsep}
\begin{enumerate}[\scriptsize a),noitemsep]
\item let $W_{res} \in \{0\}^{ d \times d}$\;
\item $\vec{r}=\vec{x}-D_{\xi}(\vec{x})$\;
\item $\vec{b}_{res} \leftarrow \vec{r}$\;
\item define $identity: \vec{a} \mapsto \vec{a}$\;
\item $R \leftarrow generateNeuralNetwork(\text{weights}=\{W_{res}\},$\\$\hspace{168pt}\text{biases}=\{\vec{b}_{res}\},$\\$\hspace{168pt}\text{activation function}=identity)$\;
\end{enumerate}
\item define $DM_{\xi} : \vec{h} \mapsto M (D_{\xi}(\vec{h}) + R(\vec{h}))$ :
\setlist{nolistsep}
\begin{enumerate}[\scriptsize a),noitemsep]
\item $DM_{\xi} \leftarrow$ stackTogether($D_{\xi},R,M$)\;
\end{enumerate}
\item $\vec{u} \leftarrow RP(DM_{\xi}, \vec{h},\vec{y})$\;
\item return $\vec{u}$\;
\end{enumerate}

 \caption{GMLF approach in case of VAE auto-encoder}
 \label{algo:VAE}
\end{algorithm}

\section{Experimental assessment}
\label{sec:exp}
In this section, we describe the chosen experimental setup. The goal is to examine the applicability of our approach for different types of MLFs obtained by different encoders.
As stated in Section \ref{sec:method:general_description}, three different types of MLFs are evaluated:  flat (non hierarchical) segmentation, hierarchical segmentation and VAE latent coding. 
For non-hierarchical/hierarchical MLF approaches, the segmentation algorithm proposed in \cite{guimaraes2012hierarchical} was used to make MLFs, since its segmentation constraints respect the causality and the location principles reported in Section \ref{sec:method:hierarchical}. However, for the non-hierarchical method, any segmentation algorithm can be used (see for example \cite{apicella2021general}). 

For the Variational Auto-Encoder (VAE) based GMLF approach, we used a $\beta$-VAE \cite{higgins2017bvae} as MLFs builder, since it results particularly suitable for generating interpretable representations.
In all the cases, we used as image classifier a VGG16 \cite{simonyan2015very} network pre-trained on ImageNet.
MLF relevances are computed with the LRP algorithm using the $\alpha-\beta$ rule\cite{bach2015}.

In Section \ref{sec:results} we show a set of possible explanations of the classifier outputs on image sampled from STL-10 dataset \cite{coates2011analysis} and the Aberdeen data set from University of Stirling (http://pics.psych.stir.ac.uk). 
The STL10 data-set is composed of images belonging to 10 different classes (airplane, bird, car, cat, deer, dog, horse, monkey, ship, truck), and the Aberdeen database is composed of images belonging to 2 different classes (Male, Female). 
Only for the Aberdeen data-set the classifier was fine-tuned using an subset of the whole data-set as training set.

\subsection{Flat Segmentation approach}
For the flat (non-hierarchical) segmentation approach, images from the STL-10 
and the Aberdeen 
data sets are used to generate the classifier outputs and corresponding explanations. 
For each test image, a set of segments (or superpixels) $S$ are generated using the image segmentation algorithm proposed \cite{guimaraes2012hierarchical} considering just one level. 
Therefore, a one-layer neural network decoder as described in Section \ref{sec:method:hierarchical} was constructed using the segmentation $S$. 
The resulting decoder is stacked on the top of the VGG16 model and fed with the "1" vector (see figure \ref{fig:arch_hier}).
The relevance of each superpixel/segment was then computed using the LRP algorithm.

\subsection{Hierarchical Image Segmentation Approach}
As for the non-hierarchical segmentation approach, the segmentation algorithm proposed in \cite{guimaraes2012hierarchical} was used, but in this case, three hierarchically organised levels were considered.  
Thus, for each test image, 3 different sets of segments (or superpixels) $\{S_i\}_{i=1}^3$ related between them in a hierarchical fashion are generated, going from the coarsest ($i=1$) to the finest ($i=3$) segmentation level. 
Next, a hierarchical decoder is made as described in section \ref{sec:method:hierarchical} and stacked on the classifier (see Figure \ref{fig:arch_hier}). As for the non-hierarchical case, the decoder is then fed with the "1"s vector. Finally, LRP is used to obtain hierarchical explanations as follows:
1) first, at the coarsest level $i=1$, the most relevant segment $\vec{s_i}_{max}$ is selected; 
2) then, for each finer level $i>1$, the segment $\vec{s_{i}}_{max}$ corresponding to the most relevant segment belonging to $\vec{s_{i-1}}_{max}$ is chosen.

\subsection{Variational auto-encoders}
Images from the Aberdeen dataset are used to construct an explanation based on VAE encoding latent variables relevances. The VAE model was trained on an Aberdeen subset using the architecture suggested in \cite{higgins2017bvae} for the CelebA dataset. Then, an encoding of $10$ latent variables is made using the encoder network for each test image. The resulting encodings were fed to the decoder network stacked on top of the trained VGG16. Next, the LRP algorithm was applied on the decoder top layer to compute the relevance of each latent variable.

\section{Results}

\label{sec:results}
In this section we report the evaluation assessment of the different realisation of the GMLF framework described in the previous section. For the evaluation we show both qualitative and quantitative results (see Section \ref{sec:quantitative-eval}). In particular, in the first part of this section we report some examples of explanations obtained using flat and hierarchical segmentation-based MLFs, and VAE-based MLFs. Thereafter, we show an example of explanation using different types of MLFs. Finally, in Section
\ref{sec:quantitative-eval} we report a quantitative evaluation of the obtained results.

\subsection{Flat Segmentation}
\begin{figure}
    \centering
    \begin{minipage}{0.49\linewidth}
    \includegraphics[width=\linewidth, frame]{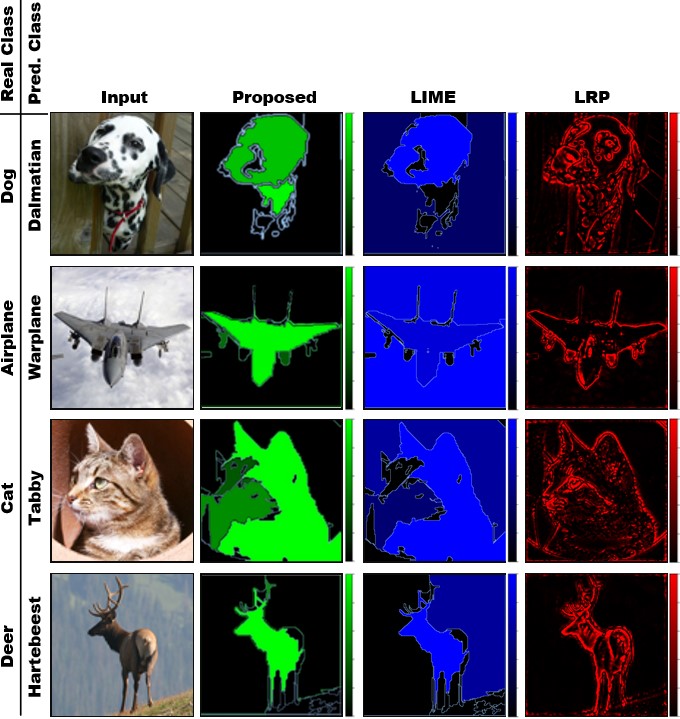}
    \captionof*{figure}{(a) STL10}
    \end{minipage}
    \begin{minipage}{0.49\linewidth}
    \includegraphics[width=\linewidth, frame]{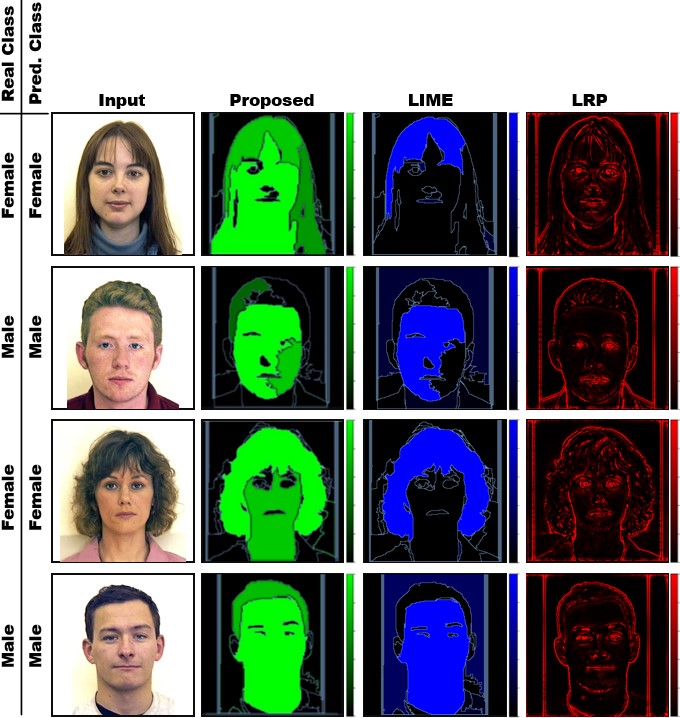}
    \captionof*{figure}{(b) Aberdeen}
    \end{minipage}
    \caption{Explanations obtained by GMLF using the flat strategy (second columns), LIME (third columns) and LRP (fourth columns) for VGG16 network responses using images from STL10 (a) and Aberdeen datasets (b). In both (a) and (b), for each input (first columns) the explanation in terms of most relevant segments are reported for the proposed flat approach (second columns) and LIME (third columns). For better clarity, we report a colormap where only the first two most relevant segments are highlighted both for MLRF and LIME.}
    \label{fig:res_flat_mlr}
\end{figure}

In Figure \ref{fig:res_flat_mlr} we show some of the explanations produced for a set of images using the flat (non hierarchical) segmentation-based experimental setup described in Section \ref{sec:method:hierarchical}. 
The proposed explanations are reported considering the first two more relevant segments according to the method described in Section \ref{sec:method:hierarchical}. 
For each image, the real class and the assigned class are reported. 
From a qualitative visual inspection, one can observe that the selected segments seem to play a relevant role for distinguishing the classes.

\subsection{Hierarchical Image Segmentation}
\label{sec:results_hierarchical}
\begin{figure}
    \centering
    \begin{minipage}{0.48\linewidth}
    \includegraphics[width=0.75\textwidth]{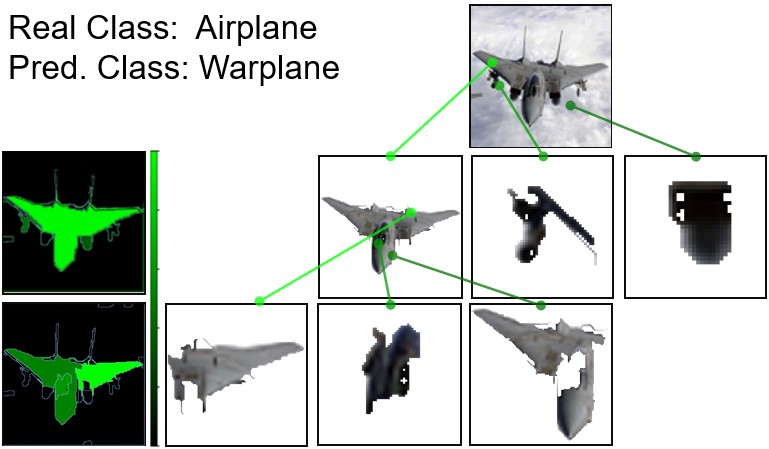}
    \end{minipage}
    \begin{minipage}{0.48\linewidth}
    \includegraphics[width=0.6\textwidth]{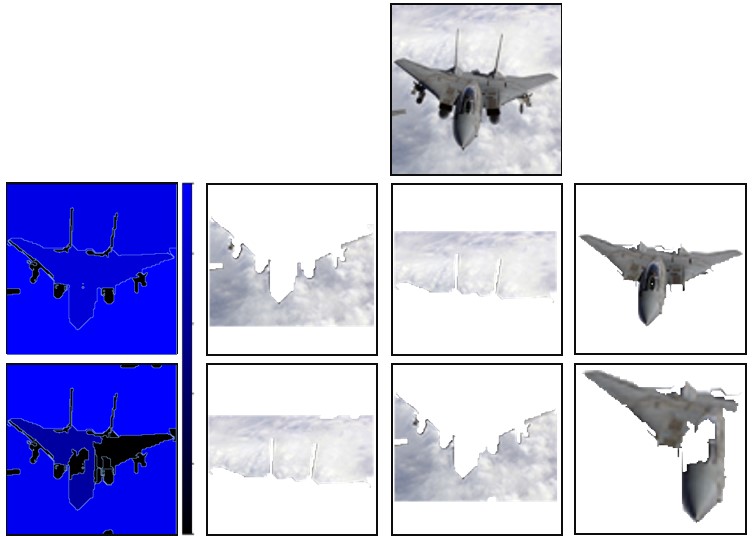}
    \end{minipage}
    \begin{minipage}{0.48\linewidth}
    \includegraphics[width=0.65\textwidth]{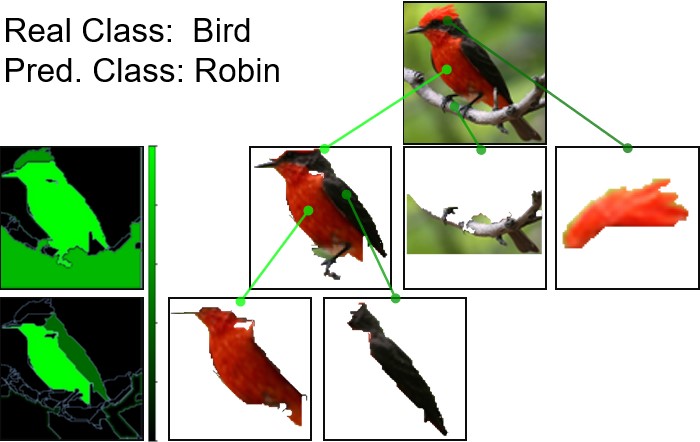}
    \end{minipage}
    \begin{minipage}{0.48\linewidth}
    \includegraphics[width=0.6\textwidth]{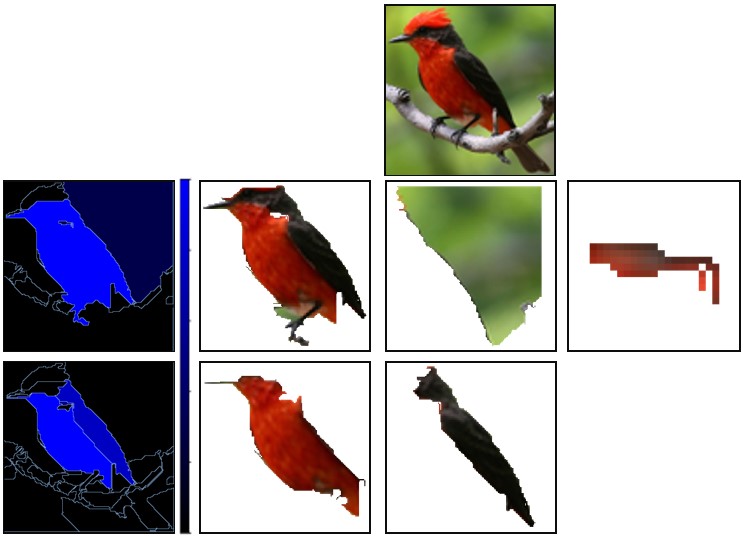}
    \end{minipage}
    \begin{minipage}{0.48\linewidth}
    \includegraphics[width=1.0\textwidth]{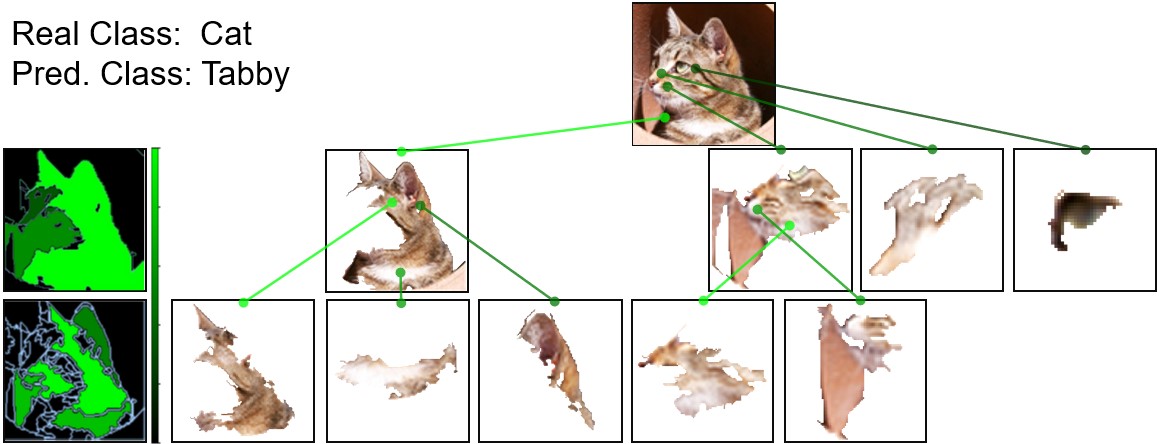}
    \end{minipage}
    \begin{minipage}{0.48\linewidth}
    \includegraphics[width=0.8\textwidth]{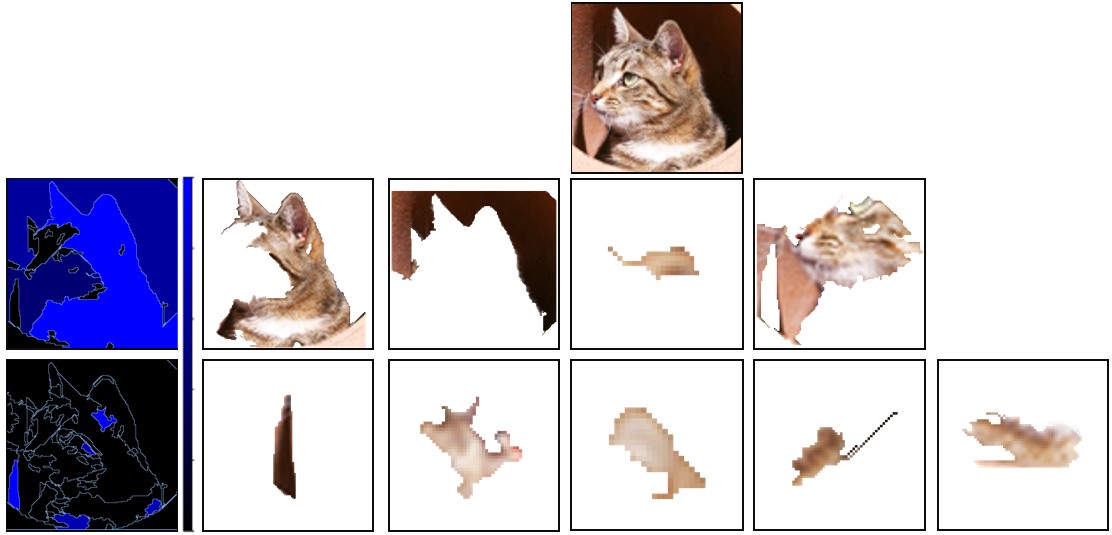}
    \end{minipage}
    \begin{minipage}{0.48\linewidth}
    \includegraphics[width=0.65\textwidth]{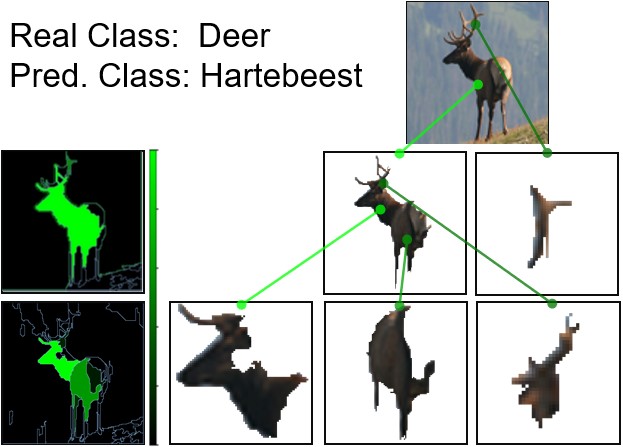}
    \end{minipage}
    \begin{minipage}{0.48\linewidth}
    \includegraphics[width=0.65\textwidth]{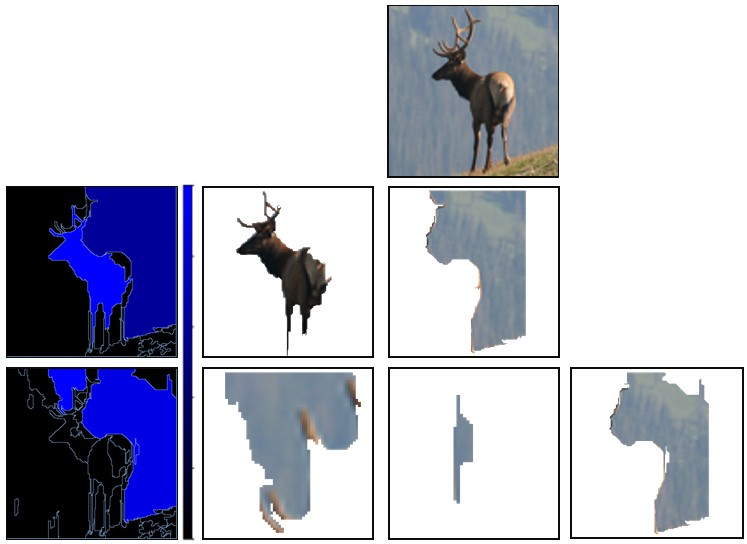}
    \end{minipage}
    \begin{minipage}{0.48\linewidth}
    \includegraphics[width=\textwidth]{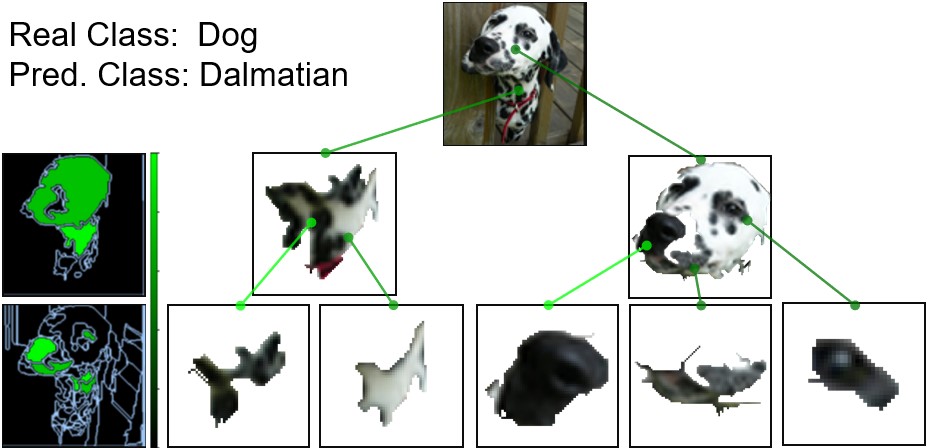}
    \captionof*{figure}{(a) Proposed approach}
    \end{minipage}
    \begin{minipage}{0.49\linewidth}
    \includegraphics[width=\textwidth]{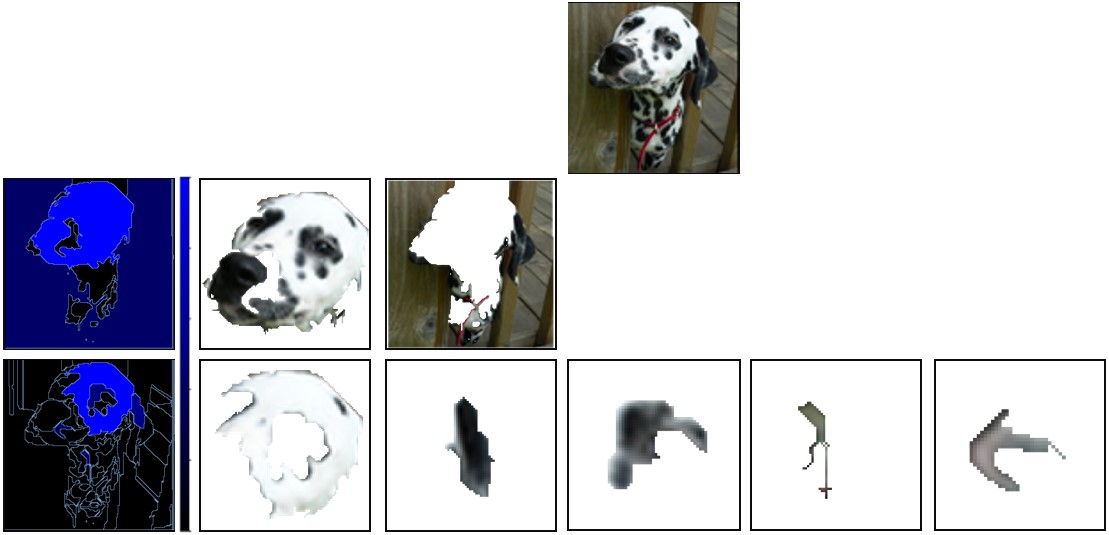}
    \captionof*{figure}{(b) LIME}
    \end{minipage}
    \caption{Examples of a two-layer hierarchical explanation on images classified as \textit{warplane, tobby, hartebeest, dalmatian} respectively by VGG16. (a) First column: segment heat map. Left to right: segments sorted in descending relevance order. Top-down: the coarsest (second row) and the finest (third row) hierarchical level. (b) LIME explanation: same input, same segmentation used in (a).}
    \label{fig:examples_hierarchical_stl10}
\end{figure}

\begin{figure}
    \centering
    \begin{minipage}{0.48\linewidth}
    \includegraphics[width=\textwidth]{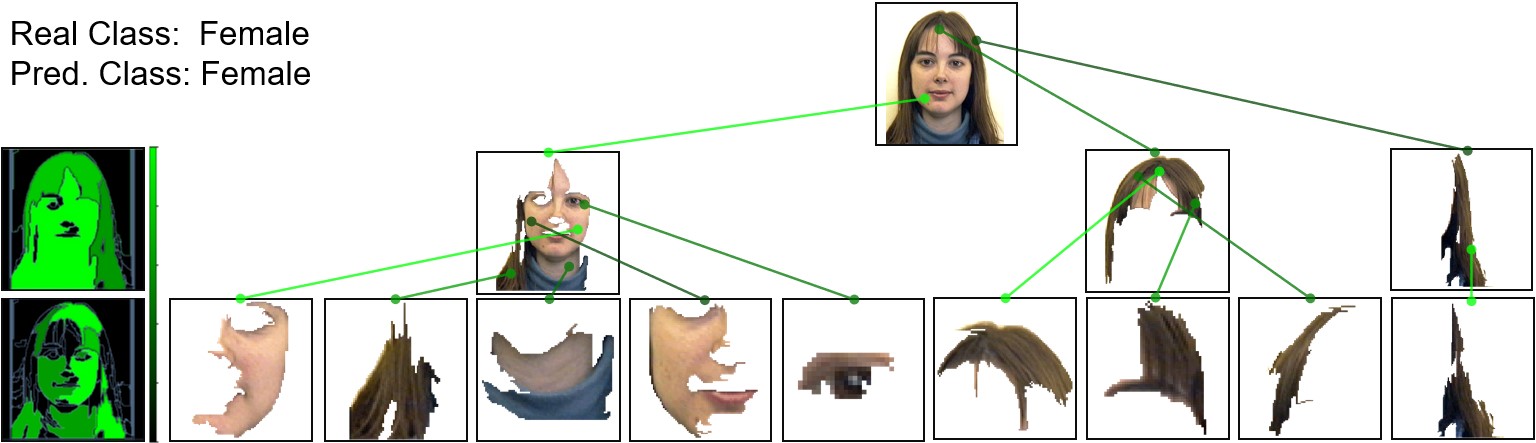}
    \end{minipage}
    \begin{minipage}{0.48\linewidth}
    \includegraphics[width=\textwidth]{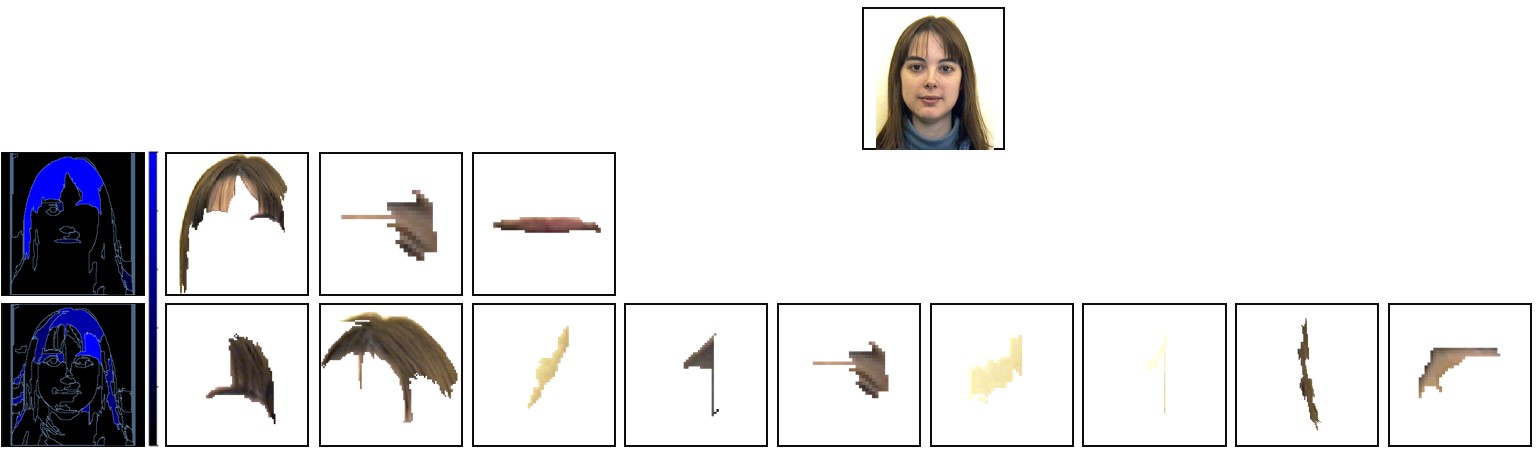}
    \end{minipage}
    
    \begin{minipage}{0.48\linewidth}
    \includegraphics[width=\textwidth]{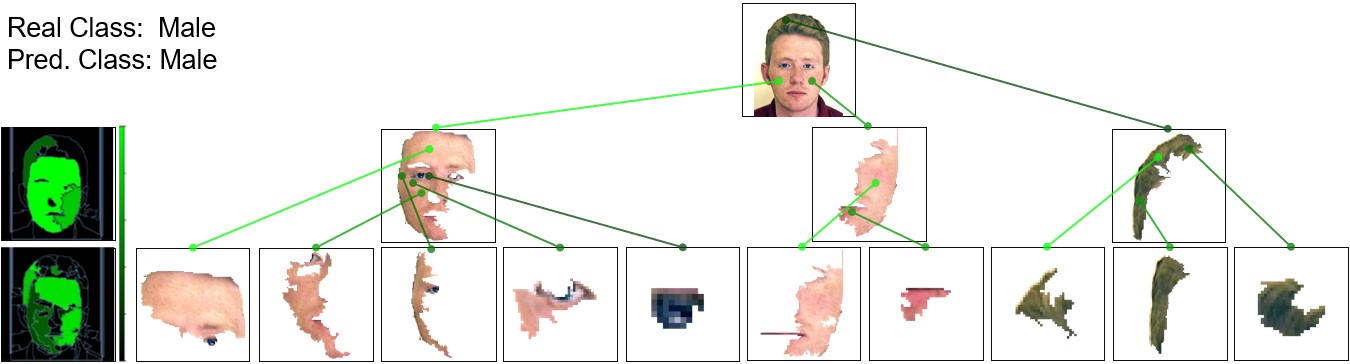}
    \end{minipage}
    \begin{minipage}{0.48\linewidth}
    \includegraphics[width=\textwidth]{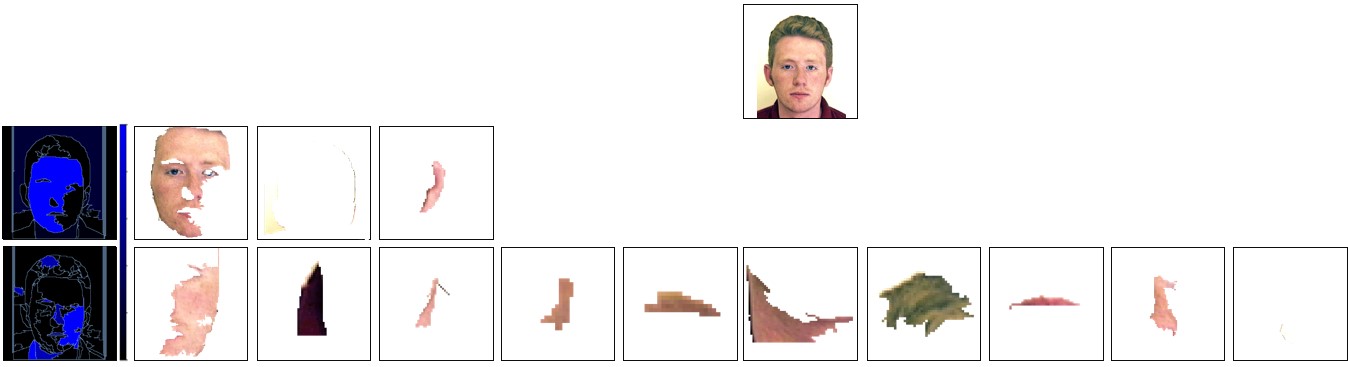}
    \end{minipage}
    
    \begin{minipage}{0.48\linewidth}
    \includegraphics[width=\textwidth]{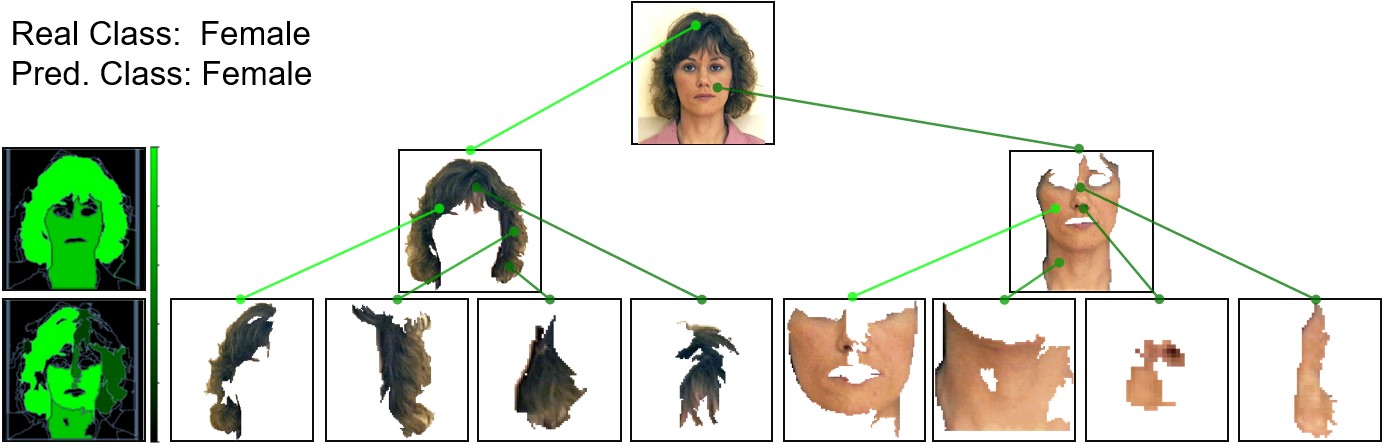}
    \end{minipage}
    \begin{minipage}{0.48\linewidth}
    \includegraphics[width=\textwidth]{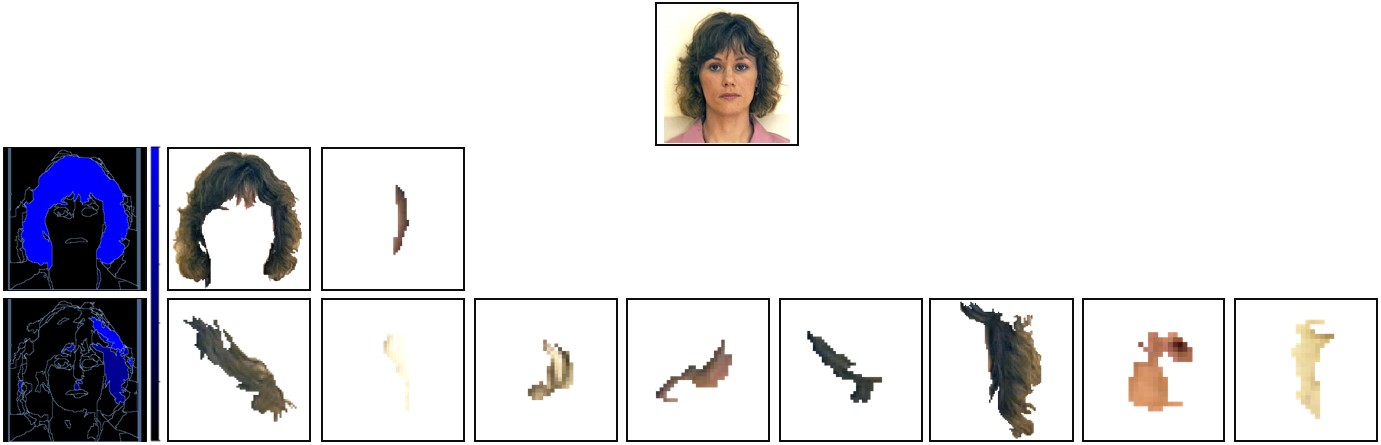}
    \end{minipage}
    
    \begin{minipage}{0.48\linewidth}
    \includegraphics[width=\textwidth]{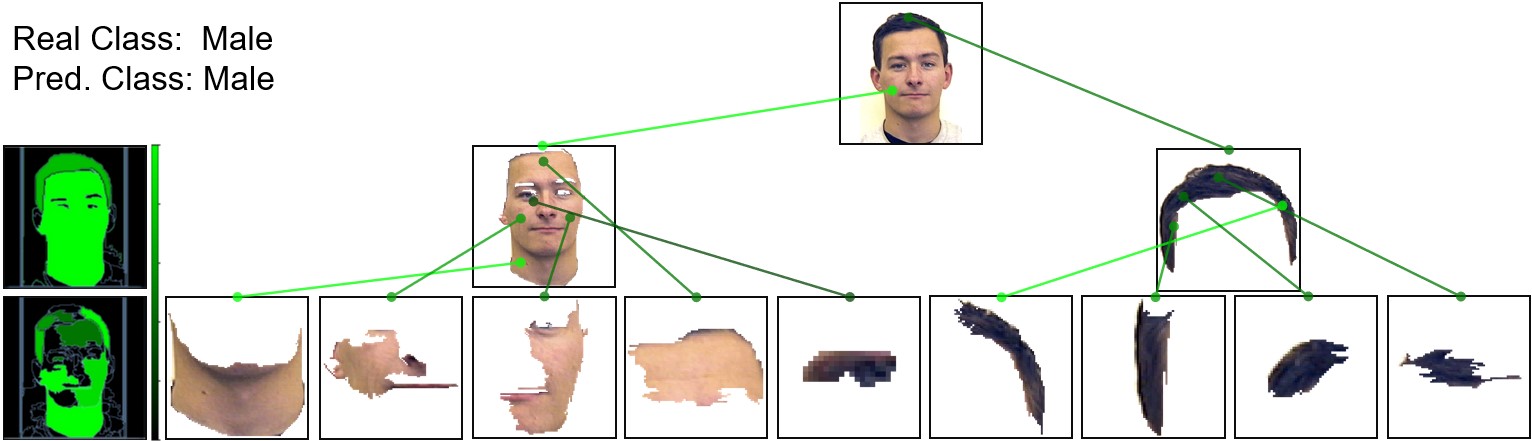}
    \captionof*{figure}{(a) Proposed approach}
    \end{minipage}
    \begin{minipage}{0.48\linewidth}
    \includegraphics[width=\textwidth]{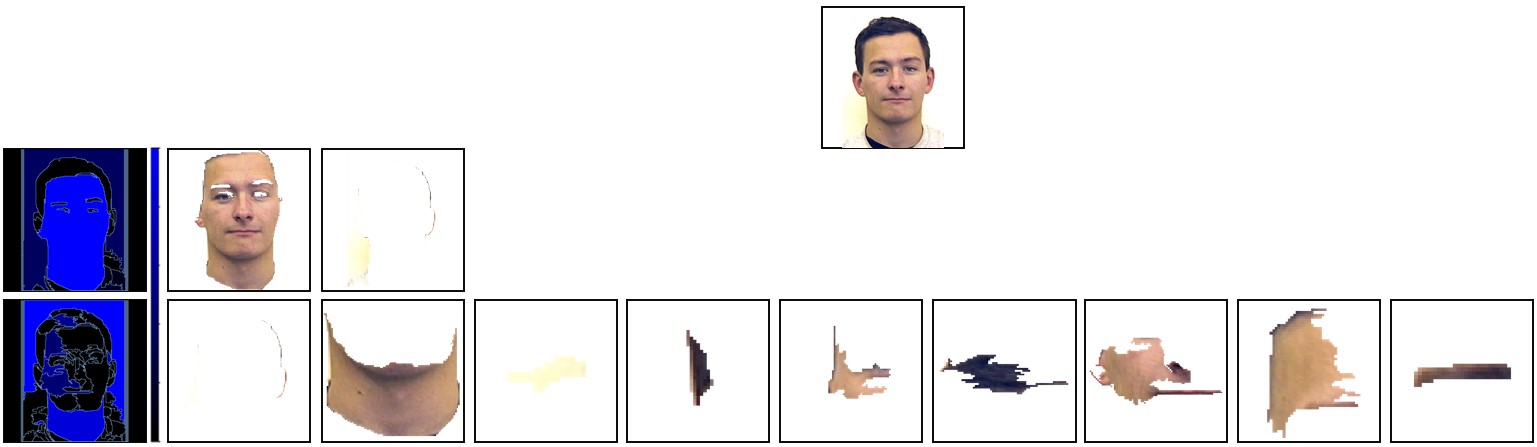}
    \captionof*{figure}{(b) LIME}
    \end{minipage}
    \caption{Examples of a two-layer hierarchical explanation on images classified as \textit{Female} and \textit{Male} by VGG16. (a) First column: segment heat map. Left to right: segments sorted in descending relevance order. Top-down: the coarsest (second row) and the finest (third row) hierarchical level. (b) LIME explanation: same input, same segmentation used in (a).}
    \label{fig:examples_hierarchical_aberdeen}
\end{figure}

In figures \ref{fig:examples_hierarchical_stl10} and \ref{fig:examples_hierarchical_aberdeen} we show a set of explanations using the hierarchical approach described in Section \ref{sec:method:hierarchical} on images of the STL10 and the Aberdeen datasets. 
In this case, we exploit the hierarchical segmentation organisation to provide MLF explanations. 
In particular, for each image, a three layers decoder has been used,  obtaining three different image segmentations $S_1$, c$S_2$ and $S_3$, from the coarsest to the finest one, which are hierarchically organised (see Section \ref{sec:method:hierarchical}). 
For the coarsest segmentation ($S_1$), the two most relevant segments $s^1_1$ and  $s^1_2$ are highlighted in the central row. 
For the image segmentation $S_2$ the most relevant segment $s^2_1$ belonging to $s^1_1$ and the most relevant segment $s^2_2$ belonging to $s^{1}_2$ are highlighted in the upper and the lower row (second column). 
The same process is made for the image segmentation $S_3$, where the most relevant segment $s^3_1$ belonging to $s^2_1$ and the most relevant segment $s^3_2$ belonging to $s^2_2$ are shown in the third column.
From a qualitative perspective, one can note that the proposed approach seems to select relevant segments for distinguishing the classes. Furthermore, the hierarchical organisation provides more clear insights about the input image's parts, contributing to the classifier decision.

\begin{figure}
    \centering
    \begin{minipage}{0.51\linewidth}
    \includegraphics[width=\textwidth]{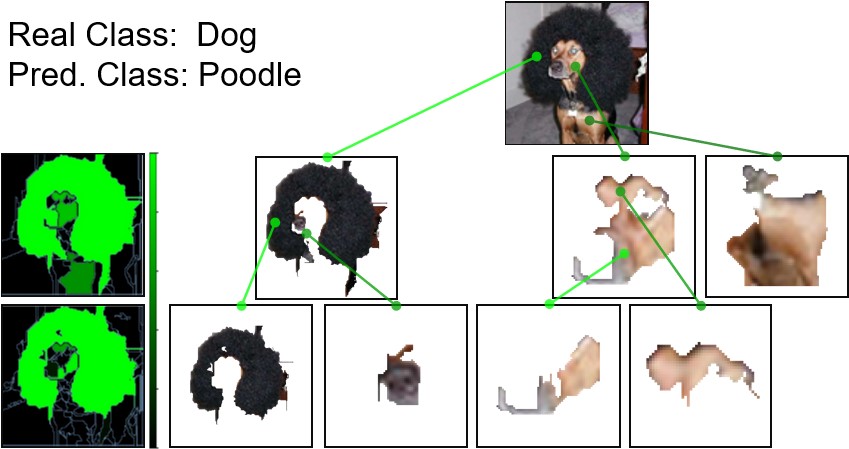}
    \captionof*{figure}{(a) classified as poodle}
    \end{minipage}
    \begin{minipage}{0.47\linewidth}
    \includegraphics[width=\textwidth]{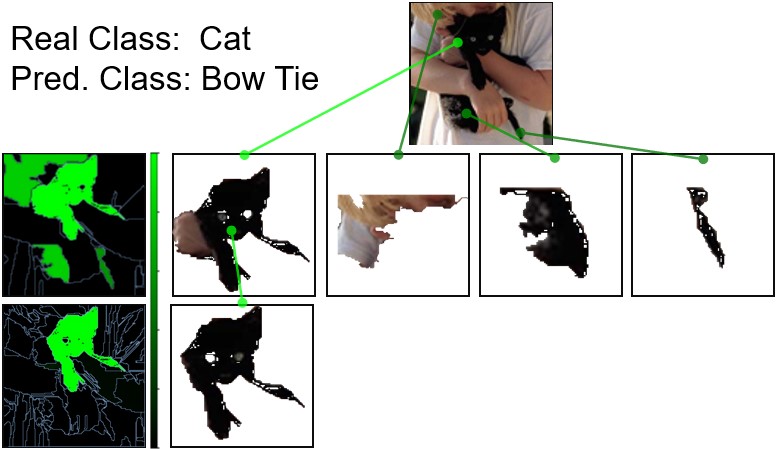}
    \captionof*{figure}{(b) classified as bow tie}
    \end{minipage}
    \caption{Results obtained by Hierarchical MLF approach (described in Section \ref{sec:method:hierarchical}) using VGG16 network on STL10 images wrongly classified by the model. (a) A dog wrongly classified as a poodle, although it is evidently of a completely different race. Inspecting the MLF explanations at different hierarchy scales, it can be seen that the classifier was probably misled by the wig (which probably led the classifier toward the poodle class), (b) A cat wrongly classified as a bow tie. Inspecting the MLF explanations at different hierarchy scales, it can be seen that the shape and the position of the cat head near the neck of the shirt, having at the same time the remaining of its body hidden, could be responsible for the wrong class.}
    \label{fig:wrong_mlr_stl10}
\end{figure}

The usefulness of a hierarchical method can also be seen in cases of wrong classifier responses. See, for example, Figure \ref{fig:wrong_mlr_stl10} where a hierarchical segmentation MLF approach was made on two images wrongly classified: 1) a dog wrongly classified as a poodle although it is evidently of a completely different race, and 2) a cat classified as a bow tie. 
Inspecting the MLF explanations at different hierarchy scales, it can be seen that, in the dog case, the classifier was misled by the wig (which probably led the classifier toward the poodle class), while, in the other case, the cat head position near the neck of the shirt, while the remaining part of the body is hidden, could be responsible for the wrong classification.

\subsection{VAE-based MLF explanations}
\begin{figure}
    \centering
    \includegraphics[width=0.30\textwidth]{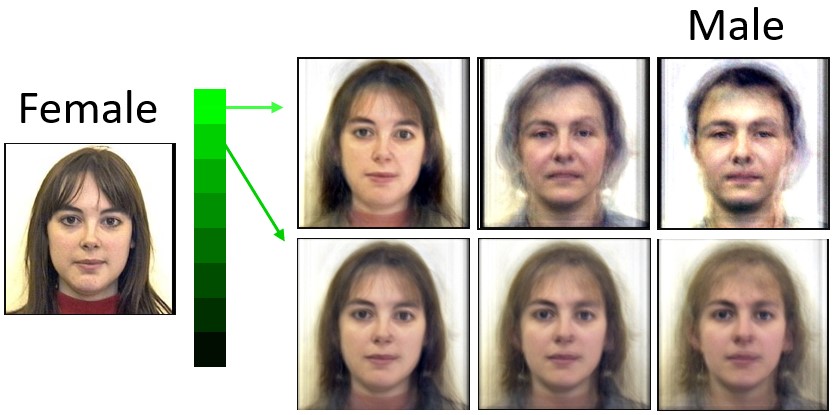}
    \hspace{0.25cm}
    \includegraphics[width=0.30\textwidth]{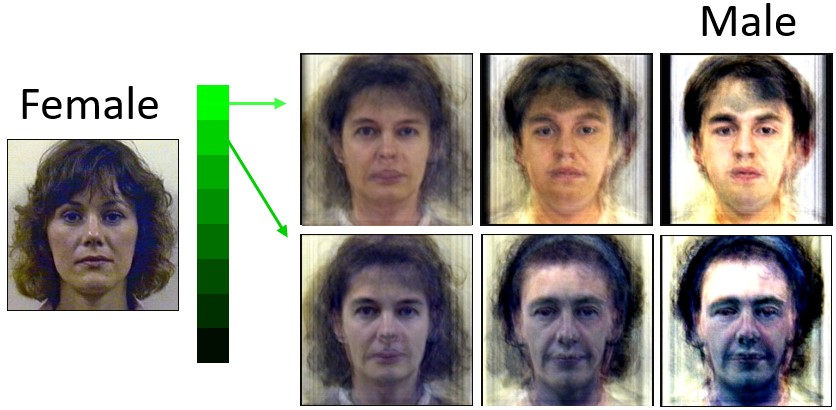}
    \hspace{0.25cm}
    \includegraphics[width=0.30\textwidth]{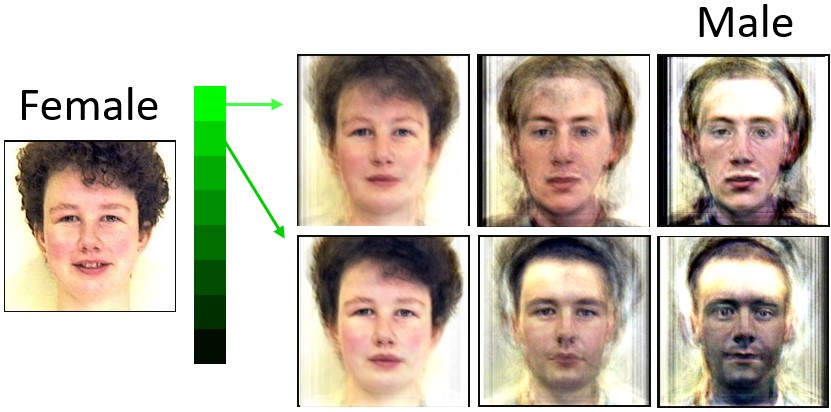}\\ 
    \vspace{0.5cm}
    \includegraphics[width=0.30\textwidth]{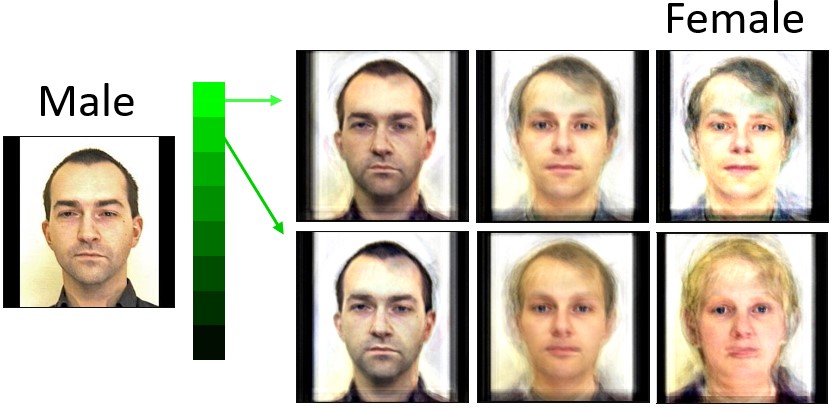}
    \hspace{0.25cm}
    \includegraphics[width=0.30\textwidth]{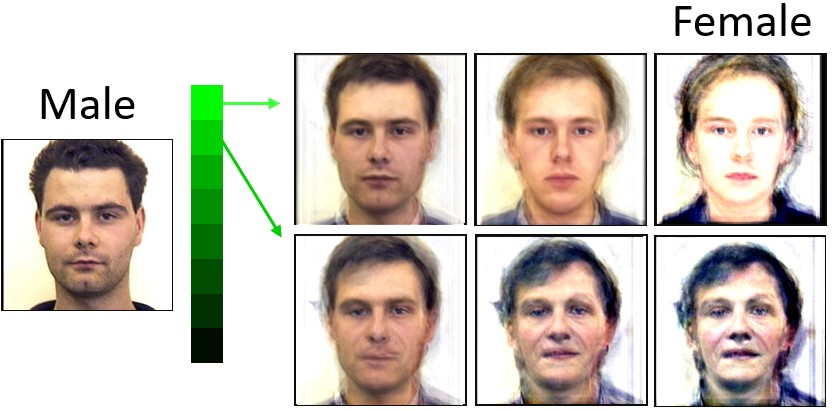}
    \hspace{0.25cm}
    \includegraphics[width=0.30\textwidth]{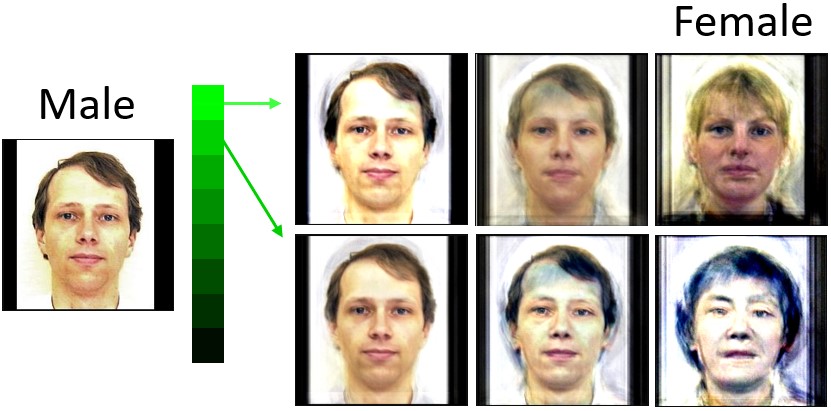}
    
    \caption{Results obtained by VAE MLF approach (described in Section \ref{sec:method:vae}) using a VGG16 network on Aberdeen image dataset. For each image, a VAE is constructed. For each input, the resulting relevance vector on the latent variable is computed. Then, decoded images are generated varying the two most relevant latent variables while fixing the other ones to the original values.}
    \label{fig:vae_mlr_abardeen}
\end{figure}

In Figure \ref{fig:vae_mlr_abardeen} a set of results using the VAE-based experimental setup described in Section \ref{sec:exp} is shown.
For each input, a relevance vector on the latent variable coding is computed. Then, a set of decoded images are generated varying the two most relevant latent variables while fixing the other ones to the original encoding values.
One can observe that varying the most relevant latent variables it seems that relevant image properties for the classifier decision are modified such as hair length and style. 

\subsection{Multiple MLF explanations}
\begin{figure}
    \centering
    \includegraphics[width=0.49\textwidth]{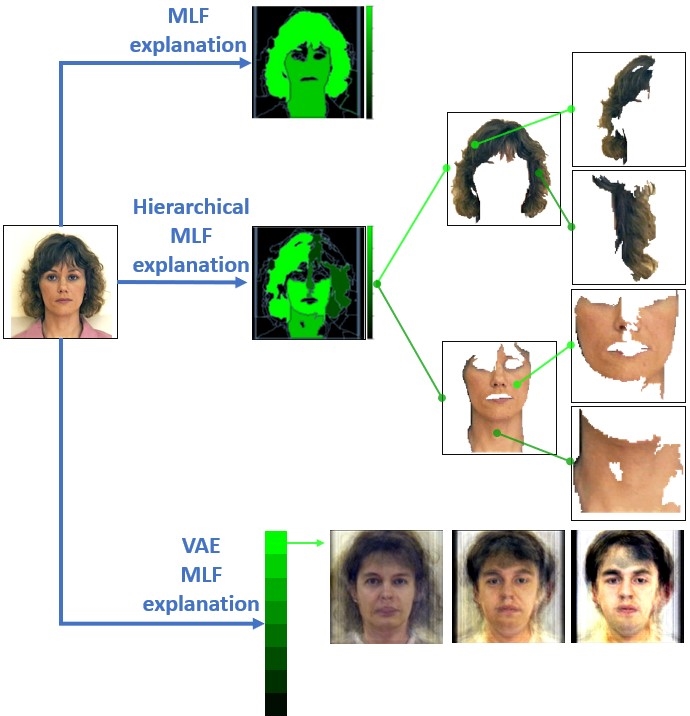}
    \includegraphics[width=0.49\textwidth]{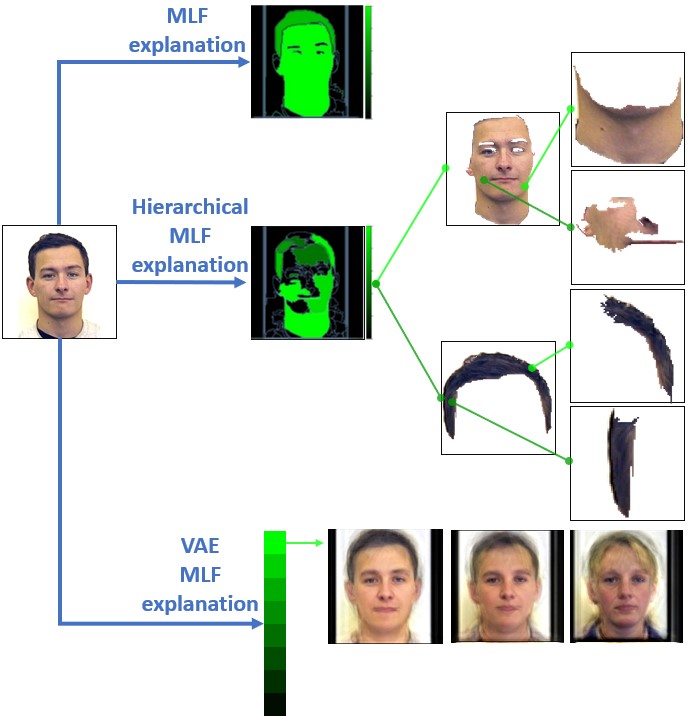}
    \caption{For each input, three different types of explanations obtained by GMLF approach are shown. In the first row, an explanation based on a flat image segmentation is reported. In the second row, an explanation based on an hierarchical segmentation. In the last row, a VAE-based MLF explanation is showed.}
    \label{fig:mlr_comparison}
\end{figure}

For the same classifier input-output, we show the possibility to provide multiple and different MLF explanations based on the three types of previously mentioned MLFs. In Figure \ref{fig:mlr_comparison}, for each input, three different types of explanations are shown. In the firs row, an explanation based on MLFs obtained by a flat image segmentation is reported. In the second row, an explanation based on MLFs obtained by an hierarchical segmentation. In the last row, a VAE-based MLF explanation is showed. Notice that the three types of explanations, although based on different MLFs, seem coherent to each other. 

\color{black}
\subsection{Quantitative evaluation}
\label{sec:quantitative-eval}
A quantitative evaluation is performed adopting the MoRF (Most Relevant First) and AOPC (Area Over Perturbation Curve) \cite{bach2015, samek2016evaluating} curve analysis. In this work, MoRF curve is computed following the \textit{region flipping} approach, a generalisation of the \textit{pixel-flipping} measure proposed in \cite{bach2015}. 
In a nutshell, given an image classification, image regions (in our case segments) are iteratively replaced by random noise and fed to the classifier, following the descending order with respect to the relevance values returned by the explanation method. In this manner, more relevant for the classification output the identified MLFs are, steepest is the curve. 
Instead, AOPC is computed as: $$AOPC=\frac{1}{L+1}<\sum\limits_{k=0}^L f(\vec{x}^{(0)})-f(\vec{x}^{(k)})>$$ where $L$ is the total number of perturbation steps, $f(\cdot)$ is the classifier output score, $\vec{x}^{(0)}$ is the original input image, $\vec{x}^{(i)}$ is the input perturbed at step $i$, and $<\cdot>$ is the average operator over a set of input images. In this manner, more relevant for the classification output the identified MLFs are, greater the AOPC value is. 

\begin{figure}
    \centering
    \begin{minipage}{0.48\linewidth}
    \includegraphics[width=\textwidth]{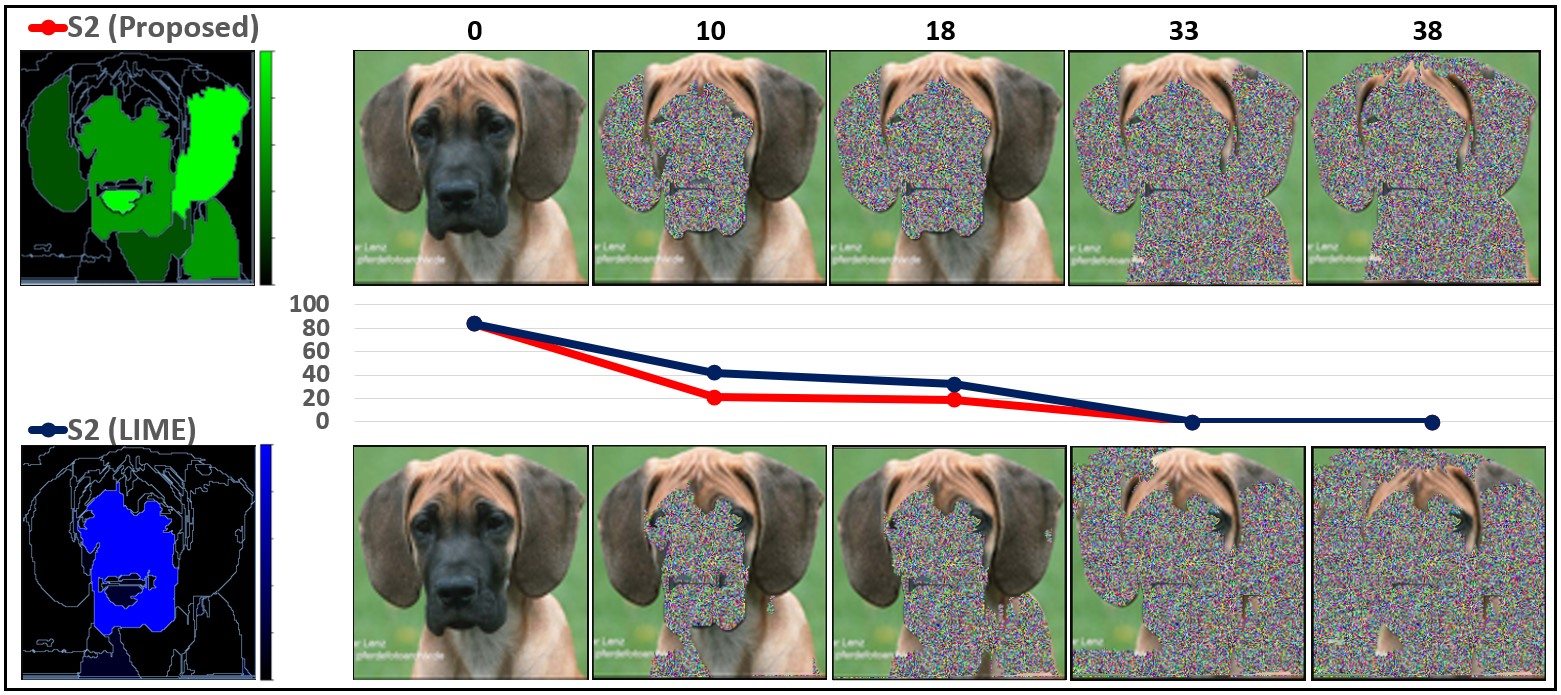}
    \includegraphics[width=\textwidth]{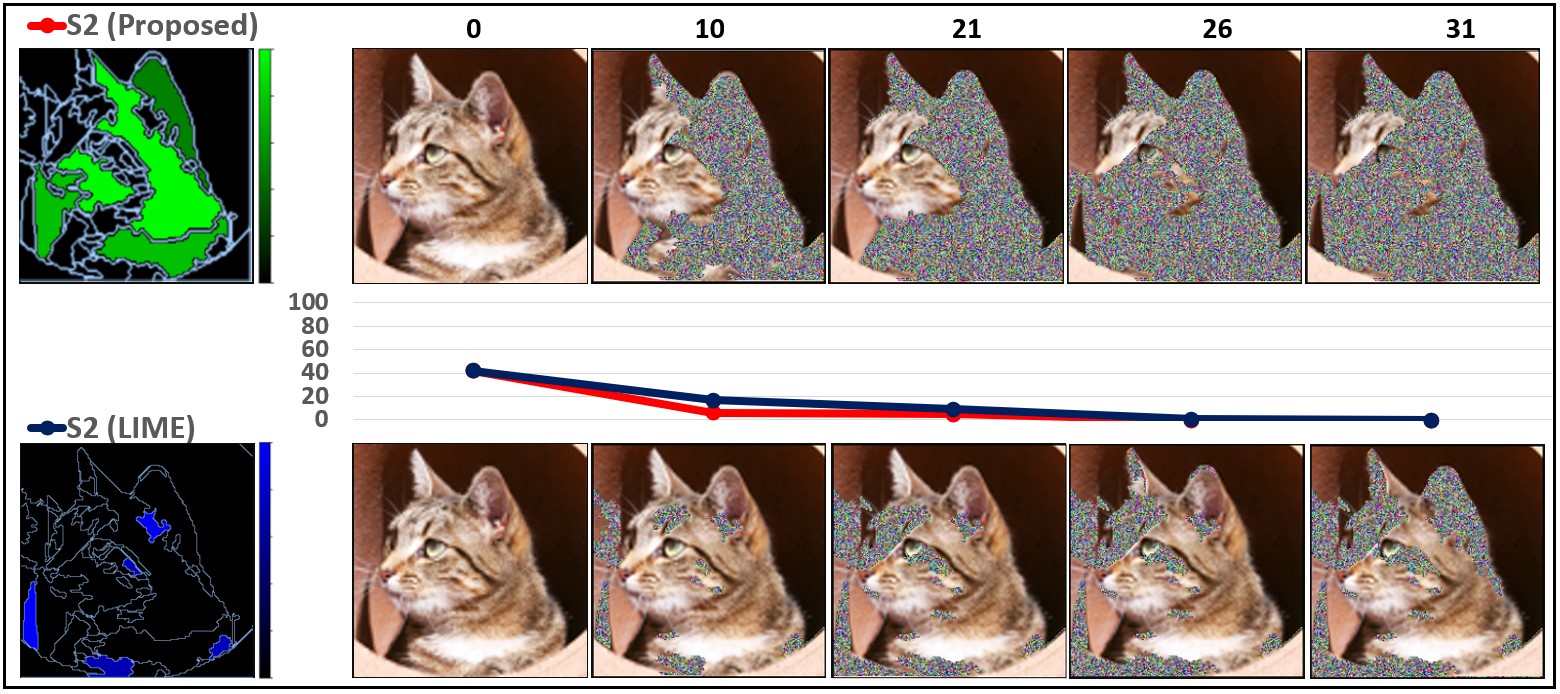}
    \captionof*{figure}{(a) STL10}
    \end{minipage}   
    \begin{minipage}{0.48\linewidth}
    \includegraphics[width=\textwidth]{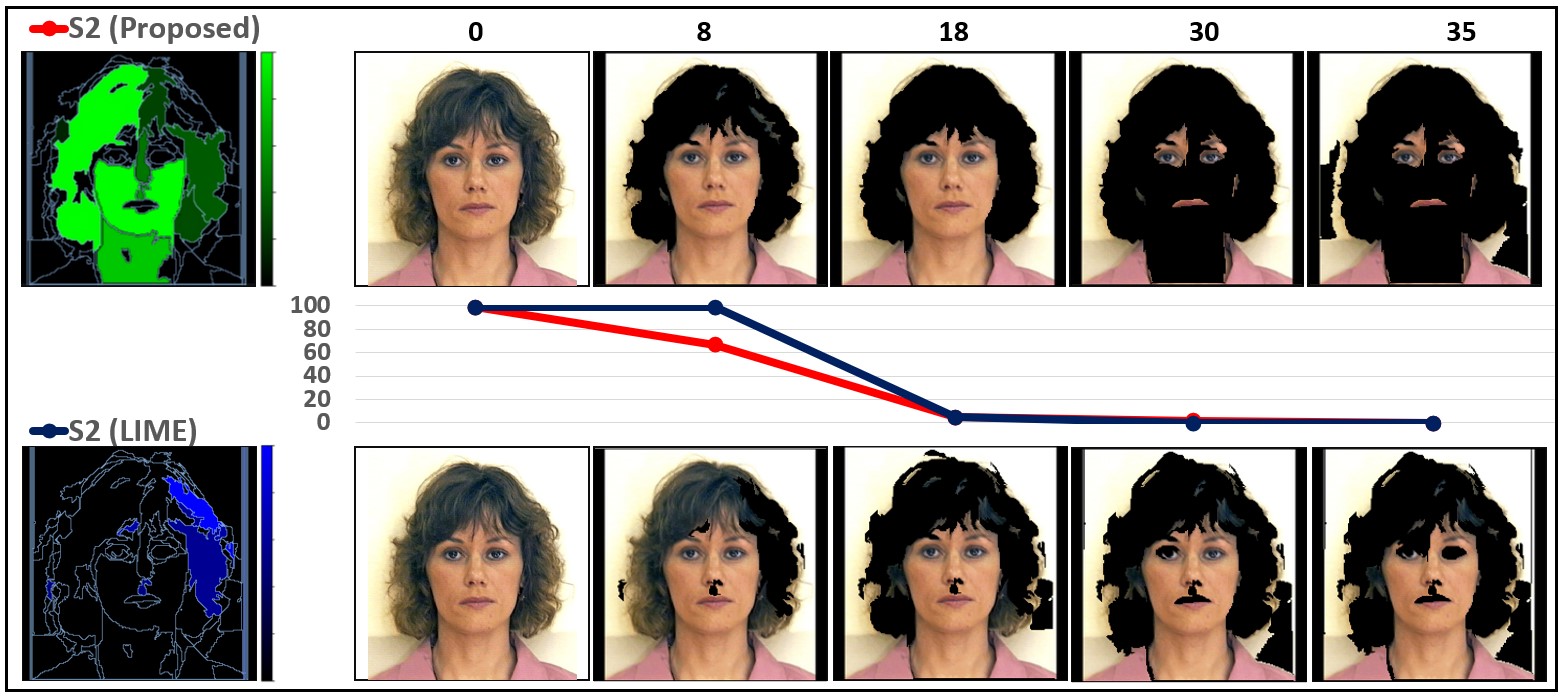}
    \includegraphics[width=\textwidth]{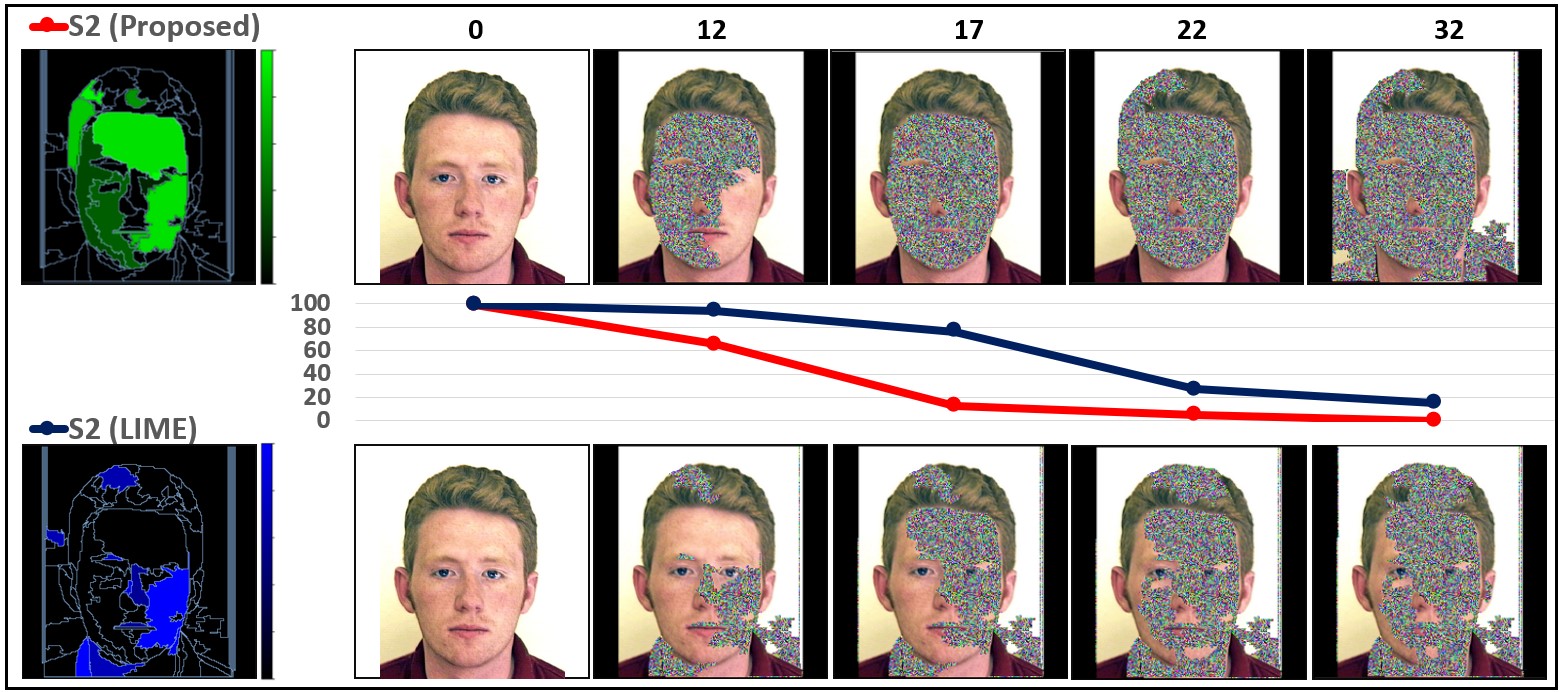}
    \captionof*{figure}{(b) Aberdeen}
    \end{minipage}
    \caption{A quantitative evaluation of the hierarchical GMLF approach on different input images. To evaluate the hierarchical GMLF approach respect to the LIME approach, a most relevant segment analysis is made using MoRF curves. MoRF curves computed with the proposed approach (red) and LIME (blue) using the last layer MLF as segmentation for both methods are shown. At each iteration step, a perturbed input based on the returned explanation is fed to the classifier. On the $y$ axis of the plot, the classification probability (in \%) of the original class for each perturbed input. On the $x$ axis, some perturbation steps. For each input image, the figures in the first and the second row show the perturbed inputs fed to the classifier at each perturbation step for the proposed explainer system and the LIME explainer, respectively. More relevant for the classification output the identified MLFs are,  steepest the MoRF curve is.}
    \label{fig:quantitative_examples_hier}
\end{figure}

\begin{figure}
    \centering
    \begin{minipage}{0.48\linewidth}
    \includegraphics[width=\textwidth]{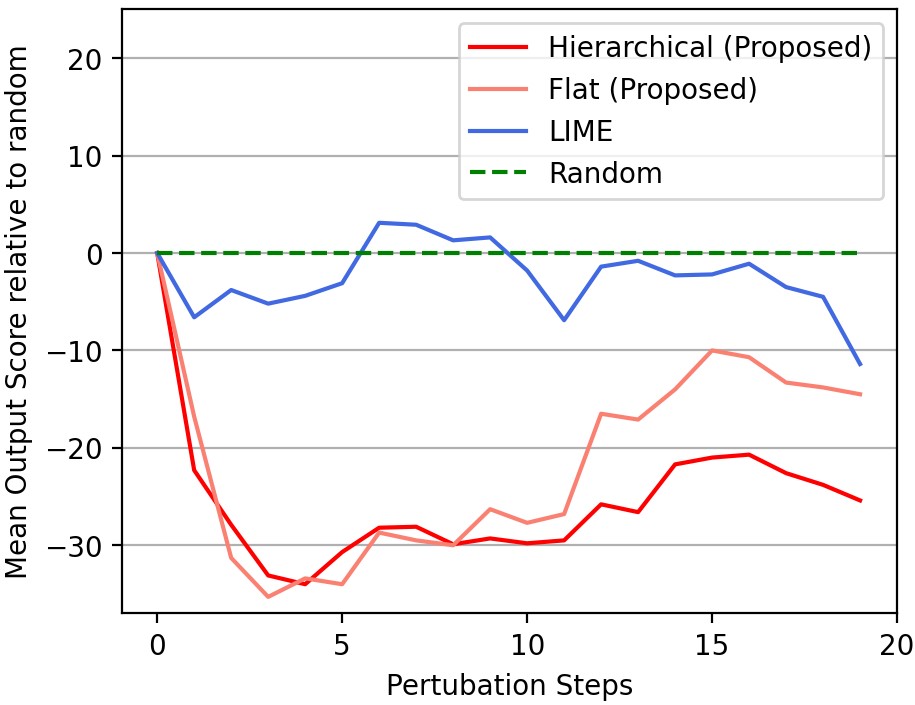}
    \includegraphics[width=\textwidth]{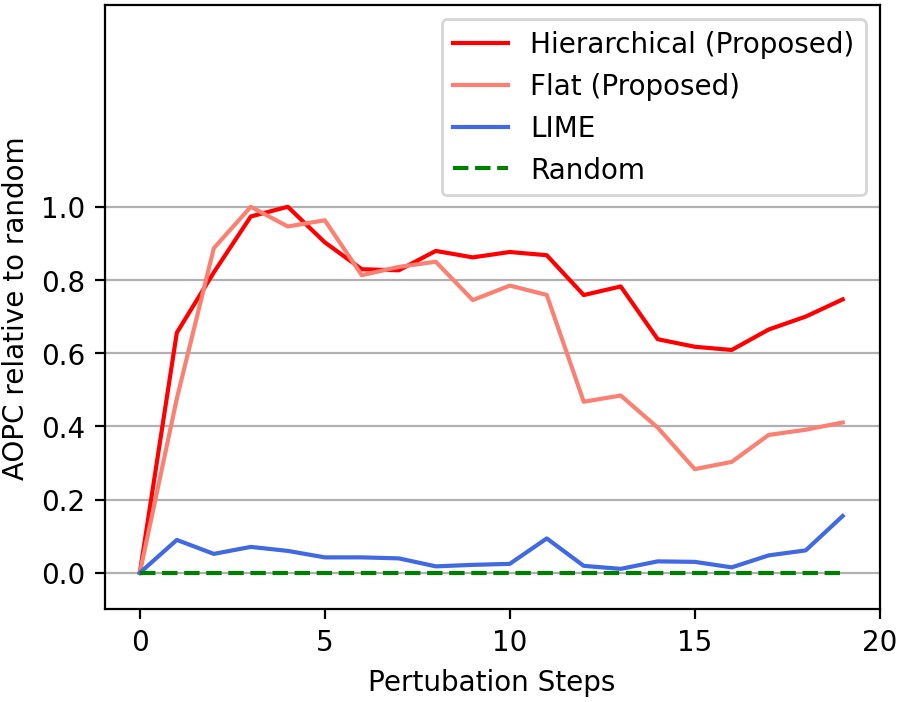}
    \captionof*{figure}{(a) STL10}
    \end{minipage}
    \begin{minipage}{0.48\linewidth}
    \includegraphics[width=\textwidth]{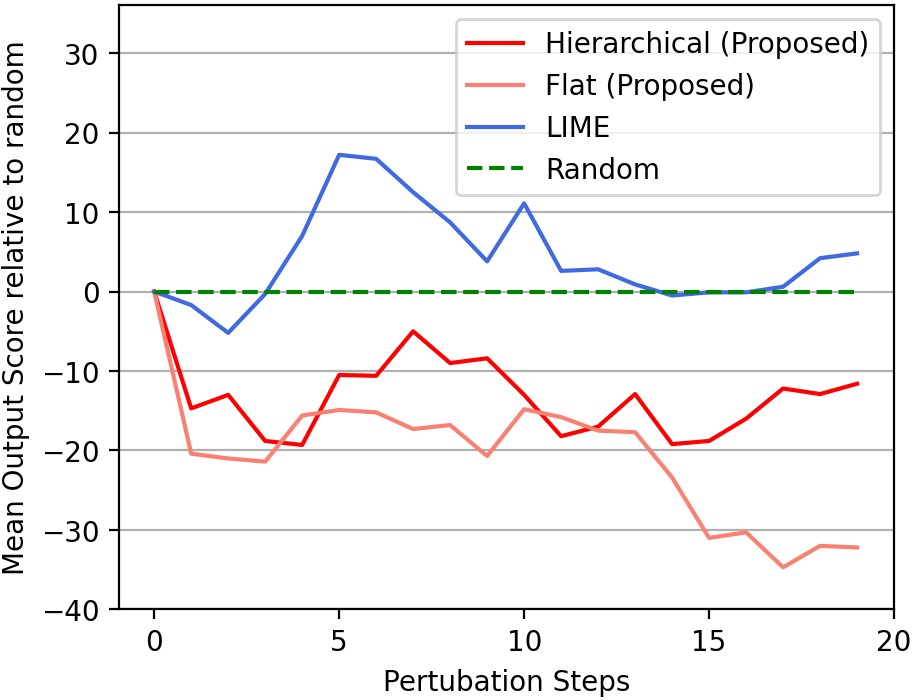}
    \includegraphics[width=\textwidth]{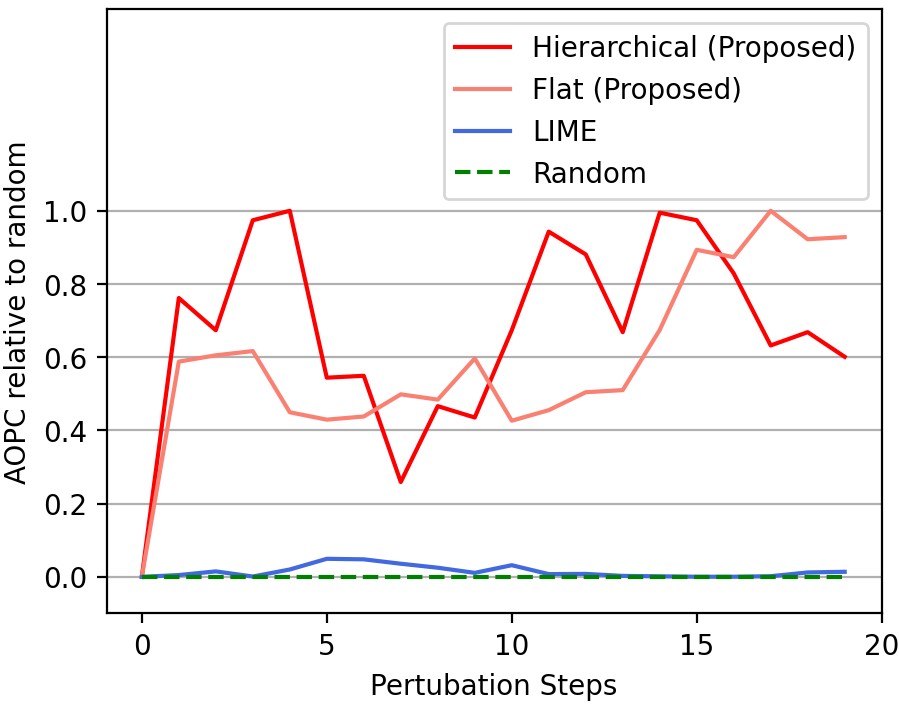}\captionof*{figure}{(b) Aberdeen}
    \end{minipage}
    \caption{average MoRF (first row) and AOPC (second row) computed on a sample of 100 and 50 random images sampled from STL10 (first column) and Abardeen (second column) respectively. Both the curves of the proposed hierarchical approach (red) and LIME (blue) are plotted using as baseline the removal of the Middle Level Features from the input images in a random order (green). More relevant for the classification output the identified MLFs are, steepest the MoRF curve is and greater the AOPC value is.
    }
    \label{fig:quantitative_mean_hier}
\end{figure}

To evaluate the hierarchical approach with respect to the flat segmentation approach, at each step, MLFs were removed from the inputs exploiting the hierarchy in a topological sort depth-first search based on the descending order's relevances. Therefore, the MLFs of the finest hierarchical layer were considered. MoRF and AOPC are shown in Fig. \ref{fig:quantitative_examples_hier} and  \ref{fig:quantitative_mean_hier}.
In Fig. \ref{fig:quantitative_examples_hier} MoRF curves for some inputs are shown. It is evident that the MLFs selected by the proposed hierarchical approach are more relevant for the produced classification output. This result is confirmed by the average MoRF and average AOPC curves (Fig. \ref{fig:quantitative_mean_hier}), obtained averaging over the MoRF and AOPC curves of a sample of $100$ and $50$ random images taken from STL10 and Aberdeen respectively. \color{black} To make an easy comparison between the proposed methods and summarising the quantitative evaluations, last iteration AOPC values of the proposed methods and LIME are reported in Tables \ref{tab:aopc_stl10} and \ref{tab:aopc_aberdeen} for STL 10 and Aberdeen dataset respectively.

\begin{figure}
    \centering
    \begin{minipage}{\linewidth}
    \includegraphics[width=\textwidth]{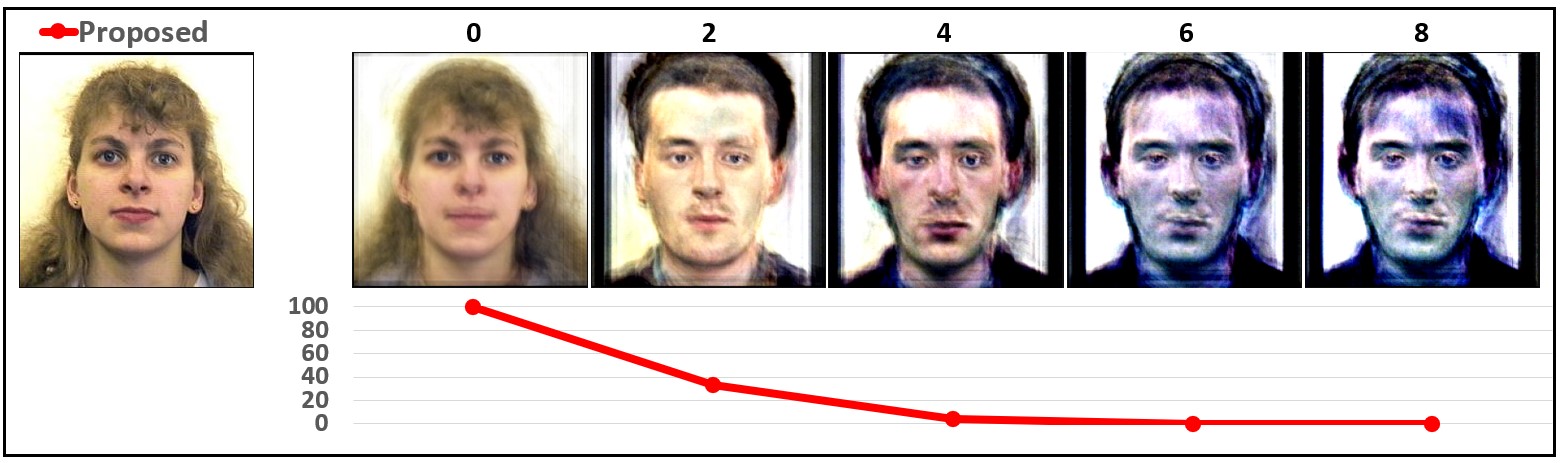}
    \includegraphics[width=\textwidth]{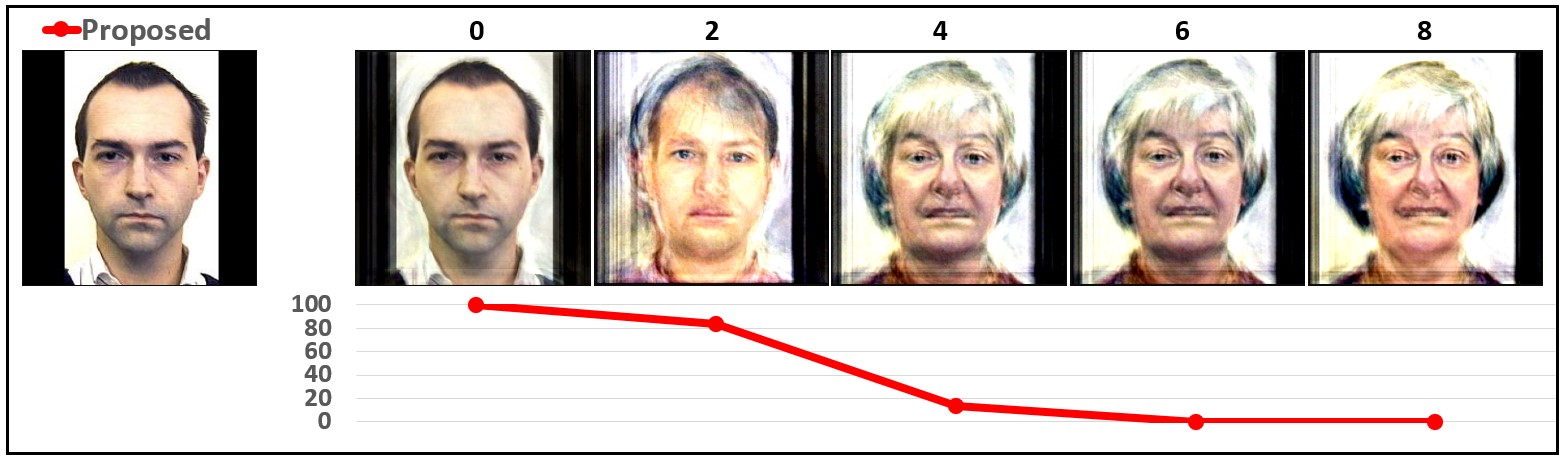}
    \captionof*{figure}{Aberdeen}
    \end{minipage}
    \caption{A quantitative evaluation of the VAE GMLF approach on different input images. MoRF curves computed with the proposed approach (red) perturbing the VAE latent variables in the order given by the explainer are shown. At each iteration step, a perturbed input based on the returned explanation is fed to the classifier. On the $y$ axis of the plot, the classification probability (in \%) of the original class for each perturbed input. On the $x$ axis, some perturbation steps. For each input image, the figures show the perturbed inputs fed to the classifier at each perturbation step for the proposed explainer system. More relevant for the classification output the identified MLFs are, steepest the MoRF curve is.}
    \label{fig:quantitative_examples_VAE}
\end{figure}

\begin{figure}
    \centering
    \begin{minipage}{0.45\linewidth}
    \includegraphics[width=\textwidth]{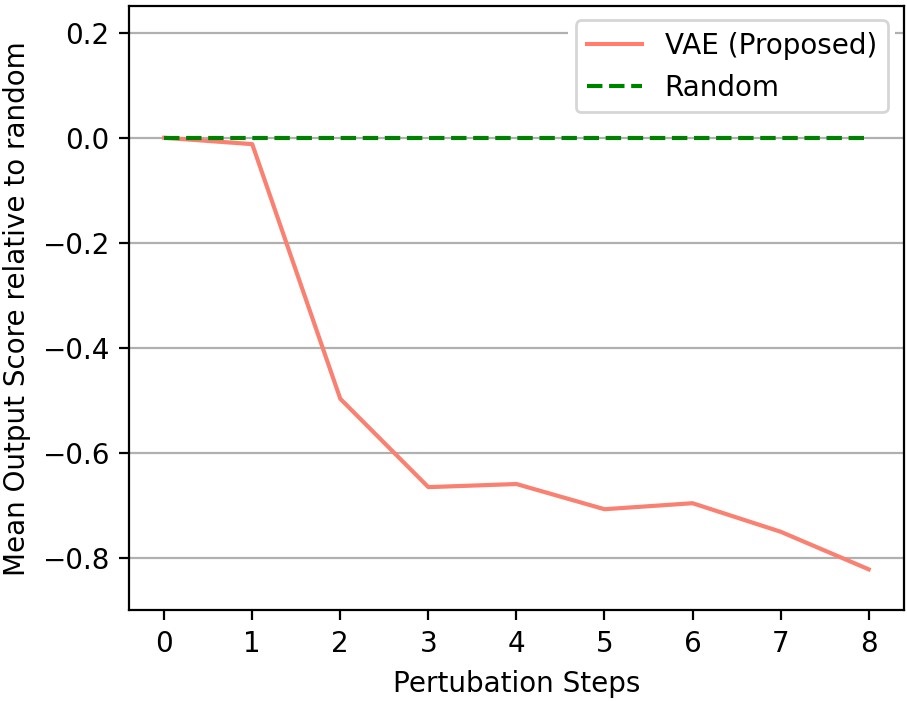}
    \captionof*{figure}{average MoRF curve}
    \end{minipage}
 \begin{minipage}{0.45\linewidth}
    \includegraphics[width=\textwidth]{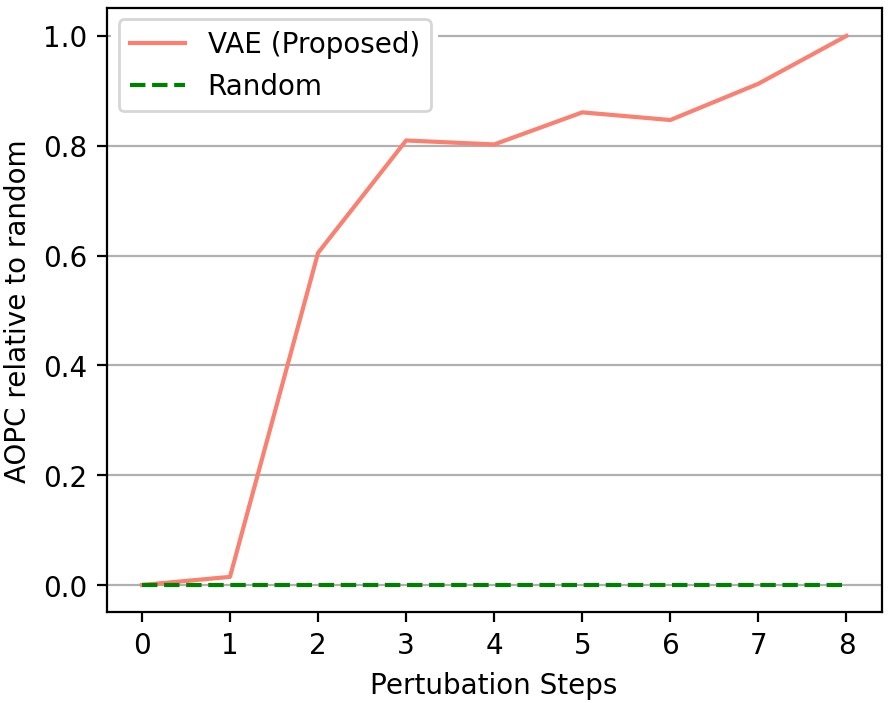}
    \captionof*{figure}{average AOPC curve}
    \end{minipage}
    \caption{average MoRF (first column) and AOPC (second column) computed on a sample of 50 random images sampled from  Aberdeen dataset. The curve proposed with the VAE approach (red) is plotted using as baseline the removal of the Middle Level Features from the input images in a random order (green). More relevant for the classification output the identified MLFs are,  steepest the MoRF curve is and greater the AOPC value is.
    }
    \label{fig:quantitative_mean_VAE}
\end{figure}

In Fig. \ref{fig:quantitative_mean_VAE}, the same quantitative analysis using the VAE strategy is shown. Examples of MoRF curves using the VAE are shown in Fig. \ref{fig:quantitative_examples_VAE}. As in the hierarchical approach, the latent features are sorted following the descending order returned by the relevance algorithm, and then noised in turn for each perturbation step. 

Due to the difference between LIME and VAE MLFs (the former corresponds to superpixels, the latter to latent variables), no comparison with LIME was reported. In our knowledge, no other study reports explanations in terms of latent variables, therefore is not easy to make a qualitative comparison with the existing methods.  Differently from perturbing the MLF of a superpixel-based approach where only an image part is substituted by noise, in a variational latent space perturbing a latent variable can lead changing in the whole input image. Therefore, classifiers fed with decoded images generated by different MLF types could return no comparable results, which may not be informative to make comparisons between MoRF curves.

\color{black}

\begin{table}[ht!]\centering
\color{black}
\caption{\color{black} average AOPC of the proposed methods and LIME obtained averaging over the last AOPC perturbation step on a sample of $100$ random images taken from STL10 dataset. Flat and hierarchical proposal are compared with LIME, resulting better in both cases. Since the LIME MLFs structure is hardly different from VAE MLFs (the former corresponds to superpixels, the latter to latent variables), the AOPC reported has not to be compared with the other results.}\label{tab:aopc_stl10}
\scriptsize
\begin{tabular}{lrr}\toprule
\textbf{} &\textbf{AOPC} \\\midrule
LIME &0.042 \\
Flat (proposed) &0.598 \\
Hierarchical (proposed) &\textbf{0.732} \\
\hdashline 
VAE (proposed) &0.595 \\
\bottomrule
\end{tabular}
\end{table}

\begin{table}[!htp]\centering
\color{black}
\caption{\color{black} average AOPC  of the proposed methods and LIME obtained averaging over the last AOPC perturbation step values on a sample of $50$ random images taken from Aberdeen dataset.}\label{tab:aopc_aberdeen}
\scriptsize
\begin{tabular}{lrr}\toprule
\textbf{} &\textbf{AOPC} \\\midrule
LIME &0.014 \\
Flat (proposed) &0.571 \\
Hierarchical (proposed) &\textbf{0.661} \\
\bottomrule
\end{tabular}
\end{table}


\section{Conclusion}
\label{sec:conclusion}
A  framework  to generate explanations in terms of middle-level features is proposed in this work. 
With the expression \textit{Middle Level Features} (MLF), (see Section \ref{sec:introduction}, we mean input features that represent more salient and understandable input properties for a user, such as parts of the input (for example, nose, ears and paw, in case of images of humans) or more abstract input properties (for example, shape, viewpoint, thickness and so on). The use of middle-level features is motivated by the need to decrease the human interpretative burden in artificial intelligence explanation systems. 

Our approach can be considered a general framework to obtain humanly understandable explanations insofar as it can be applied to different types of middle-level features as long as an encoder/decoder system is provided (for example image segmentation or latent coding) and an explanation method producing heatmaps can be applied on both the decoder and the ML system whose decision is to be explained (see Section \ref{sec:method:general_description}).
Consequently, the proposed approach enables one to obtain different types of explanations in terms of different MLFs for the same pair input/decision of an ML system, that may allow developing XAI solutions able to provide user-centred explanations according to several research directions proposed in literature \cite{ribera2019can,lim2019these}.

We experimentally tested (see Section \ref{sec:exp} and \ref{sec:results}) our approach using three different types of MLFs: flat (non hierarchical) segmentation, hierarchical segmentation and VAE latent coding. 
Two different datasets were used: STL-10 dataset 
and the Aberdeen dataset from the University of Stirling.

We evaluated our results from both a qualitative and a quantitative point of view. The quantitative evaluation was obtained using MoRF curves \cite{samek2016evaluating}. 

The results are encouraging, both under the qualitative point of view, giving easily human interpretable explanations, and the quantitative point of view, giving comparable performances to LIME. Furthermore, we show that a hierarchical approach can provide, in several cases, clear explanations about the reason behind classification behaviours.

\section*{Acknowledgment}
This work  is supported by the European Union - FSE-REACT-EU, PON Research and Innovation 2014-2020 DM1062/2021 contract number 18-I-15350-2 and by the Ministry of University and Research, PRIN research project "BRIO – BIAS, RISK, OPACITY in AI: design, verification and development of Trustworthy AI.", Project no. 2020SSKZ7R .
\bibliography{bibliography.bib}







\end{document}